\documentclass[twoside]{article}

\usepackage[accepted]{aistats2026}
\usepackage[round]{natbib}

\usepackage[utf8]{inputenc} 
\usepackage[T1]{fontenc} 
\usepackage{microtype} 
\usepackage{nicefrac}

\usepackage{enumitem}

\usepackage{titletoc}
\usepackage[page,header]{appendix}

\usepackage[hyphens]{url} 
\usepackage{hyperref}
\hypersetup{
    colorlinks,
    citecolor=blue,
    filecolor=blue,
    linkcolor=blue,
    urlcolor=blue,
}

\usepackage{graphicx}
\usepackage{chngcntr} 

\usepackage{makecell}
\usepackage{tabularx}
\usepackage{wrapfig}
\usepackage{booktabs}
\usepackage{multirow}
\usepackage{colortbl}
\usepackage{tablefootnote}
\usepackage{array}
\usepackage{tikz}

\usepackage{algorithm}
\usepackage{algorithmic}

\usepackage{amsmath}
\usepackage{amssymb}
\usepackage{amsfonts}
\usepackage{amsthm}

\usepackage{caption}
\usepackage{subcaption}
\definecolor{bright-gray}{HTML}{EEEEEE}
\definecolor{tab:pink}{HTML}{E377C2}
\definecolor{tab:purple}{HTML}{9467BD}
\definecolor{tab:cyan}{HTML}{17BECF}
\definecolor{tab:blue}{HTML}{1F77B4}

\newcommand{\pink}[1]{{\color{tab:pink} #1}}
\newcommand{\purple}[1]{{\color{tab:purple} #1}}

\newcommand{\operation}[1]{\operatorname{#1}}

\begin{document}

%
\runningtitle{Learning Hyperparameters via a Data-Emphasized Variational Objective}

%

\twocolumn[

\aistatstitle{Learning Hyperparameters via a Data-Emphasized\\Variational Objective}

\aistatsauthor{Ethan Harvey \And Mikhail Petrov \And Michael C. Hughes}

\aistatsaddress{Tufts University\\\texttt{ethan.harvey@tufts.edu} \And Tufts University\\\texttt{mikhail.petrov@tufts.edu} \And Tufts University\\\texttt{michael.hughes@tufts.edu}}

]

\begin{abstract}
When training large models on limited data, avoiding overfitting is paramount.
Common grid search or smarter search methods rely on expensive separate runs for each candidate hyperparameter, while carving out a validation set that reduces available training data. 
In this paper, we study gradient-based learning of hyperparameters via the evidence lower bound (ELBO) objective from Bayesian variational methods. This avoids the need for any validation set.
We focus on scenarios where the model is over-parameterized for flexibility and the approximate posterior is chosen to be Gaussian with isotropic covariance for tractability, even though it cannot match the true posterior.
In such scenarios, we find the ELBO prioritizes posteriors that match the prior, leading to severe underfitting.
Instead, we recommend a data-emphasized ELBO that upweights the likelihood but not the prior.
In Bayesian transfer learning of image and text classifiers, our method reduces the 88+ hour grid search of past work to under 3 hours while delivering comparable accuracy.
We further demonstrate how our approach enables efficient yet accurate approximations of Gaussian processes with learnable lengthscale kernels.
\end{abstract}

\section{Introduction}

When training deep neural networks (DNNs) or other large models, significant time and effort are devoted to avoid overfitting.
A common strategy is to use grid search to find hyperparameters that perform best on a validation set~\citep{raschka2018model}.
Smarter search strategies include successive halving~\citep{karnin2013almost}, Bayesian optimization~(BO;~\citealp{snoek2012practical,hvarfner2024vanilla}), or meta-learned BO~\citep{wang2024pre}.
Such searches have two disadvantages.
First, they require expensive separate runs for each candidate hyperparameter.
Second, they need to carve out a labeled validation set, reducing data for model training. This is worrisome when available data has limited size.

\let\thefootnote\relax\footnotetext{\hspace{-20pt} Code: \href{https://github.com/tufts-ml/data-emphasized-ELBO}{\texttt{github.com/tufts-ml/data-emphasized-ELBO}}}

A Bayesian approach to hyperparameters seems to be an elegant and pragmatic solution.
Suppose we model observations $y_{1:N}$ via a likelihood $p_{\eta}(y_{1:N} | \theta)$, where $\theta$ is a high-dimensional parameter to be estimated and $\eta$ is a hyperparameter vector. We also assume a prior $p_{\eta}(\theta)$. To learn both $\theta$ and $\eta$ from data, we can follow the type-II maximum likelihood recipe: estimate the posterior $p_{\eta}(\theta | y_{1:N})$ while simultaneously learning $\eta$ to maximize $p_{\eta}(y_{1:N}) = \int_{\theta} p_{\eta}(y_{1:N}, \theta) d\theta$. The objective $p_{\eta}(y_{1:N})$ is known as the \emph{marginal likelihood} or \emph{evidence}~\citep{mackay1996hyperparameters,neal1996bayesian,rasmussen2006bayesian}.
The marginal likelihood favors $\eta$ values that lead to simpler models that fit the data well while avoiding overfitting~\citep{jeffreys1939theory,mackay1991bayesian,rasmussen2000occam,grunwald2005tutorial}. This objective naturally penalizes overcomplexity and is often praised as a Bayesian Occam's razor.
Learning $\eta$ to maximize the marginal likelihood via gradient ascent resolves both issues raised above: we need only one run of gradient ascent (not separate runs for each candidate $\eta$) and can use all available data for training without overfitting. No validation set is needed at all.

Unfortunately, for large flexible models of practical interest, the marginal likelihood strategy appears underutilized despite being well-known for decades. 
Instead, recent works in Bayesian transfer learning~\citep{krishnan2019efficient,krishnan2020specifying,shwartz2022pre,harvey2024transfer,rudner2024finetuning} employ grid searches that take \emph{multiple days}.
One obvious barrier is evaluating the marginal likelihood. For modern DNNs, computing this high-dimensional integral is difficult even for a specific $\eta$. Learning $\eta$ via gradient ascent is tougher.
\citet{immer2021scalable} offer a route to gradient-based learning of $\eta$ via a Laplace approximation of the log marginal likelihood, yet their method's runtime is expensive due to an estimated Hessian matrix.
Alternatively, variational Bayesian methods~\citep{jordan1999introduction,blei2017variational} promise a tractable objective which lower bounds the log marginal likelihood (LML), known as the evidence lower bound (ELBO).
However, the ELBO is not widely used to learn $\eta$ for DNNs. \citet{blundell2015weight} tried to learn hyperparameters of Bayesian neural networks (BNNs) via gradients of the ELBO but found it ``to not be useful, and yield worse results,'' although no concrete comparison was provided.

In this work, we study when and why ELBO-based methods fail to provide reliable model selection for large models like DNNs.
We focus on a target scenario with two key assumptions:
\begin{itemize}[noitemsep,leftmargin=*,topsep=0pt]
    \item First, models are over-parameterized such that $D\gg N$, where $D$ is the size of parameter $\theta$ and $N$ is the number of training examples. Such models often enjoy practical success~\citep{li2018explicit} if $\eta$ is selected to avoid overfitting.
    \item Second, to stay affordable we assume the approximate posterior is Gaussian with simplified covariance. With large models, we cannot afford to estimate a $D \times D$ covariance matrix. Variational methods let us explore isotropic posteriors with $D+1$ parameters and runtime close to standard point estimation.
\end{itemize}

Our work makes two contributions for this target scenario.
First, we show analytically and empirically that the ELBO objective favors posteriors and hyperparameters that underfit the data, substantiating \citeauthor{blundell2015weight}'s claim of ``worse results''.
Second, to remedy this issue we suggest an alternative objective that we call the \emph{data-emphasized ELBO} (DE-ELBO). By upweighting the likelihood term inside the ELBO, our DE-ELBO can jointly learn posteriors and hyperparameters that fit the data better.

We pursue two case studies to justify and validate our approach.
First, Case Study A (Sec.~\ref{sec:caseA}) looks at a $D \gg N$ regression model where a true posterior with full covariance is analytically tractable. 
Here we can prove in our target scenario why the ELBO will match a prior variance and thus underfit, while our DE-ELBO delivers better data fits.
Second, Case Study B (Sec.~\ref{sec:caseB}) on DNN transfer learning for image and text classification empirically covers many datasets, model families, and architectures like ResNets, ViTs, and ConvNeXts. 
In all studies, our  data-emphasized ELBO yields far better practical fits than the standard ELBO. Moreover, our DE-ELBO yields competitive or better fits than recent Bayesian methods like~\citet{immer2021scalable} or~\citet{lotfi2022bayesian}, even when they use diagonal (not isotropic) covariances.
There are also  speed wins: our approach reduces the 88+ hour grid search of recent works~\citep{shwartz2022pre,rudner2024finetuning} to under 3 hours
while giving comparable or better accuracy.
Our approach thus advances the accuracy-runtime frontier of practical Bayesian model selection.

\section{Background}
\label{sec:background}

Our generic probabilistic model assumes observed data $\{y_i \}_{i=1}^N$ are \emph{i.i.d.} conditioned on parameters $\theta \in \mathbb{R}^D$:
\begin{align}
\label{eq:joint_pdf_template}
    p_{\eta}( y_{1:N}, \theta) = p_{\eta}(\theta ) \cdot  \textstyle \prod_{i=1}^N p_{\eta}( y_i | \theta).
\end{align}
This template is instantiated by specifying a concrete likelihood $p_{\eta}(y_i | \theta)$ and prior $p_{\eta}(\theta)$. The subscript indicates possible dependence on hyperparameters $\eta$.

Direct estimation of the posterior $p_{\eta}( \theta | y_{1:N})$ or marginal likelihood is typically intractable. Instead, we can pursue an approximate posterior via variational methods~\citep{jordan1999introduction,blei2017variational}. We first select an ``easy-to-use'' family of distributions over the parameter $\theta$. A member of this family is denoted $q_{\psi}(\theta)$, where each variational parameter $\psi$ defines a specific distribution over $\theta$. We then pose an optimization problem: find the variational parameter $\psi$ that makes $q_{\psi}(\theta)$ as close as possible to the true (intractable) posterior.

We can tractably estimate $\psi$ by maximizing the \emph{evidence lower bound} (ELBO; \citealp{blei2017variational}), defined for our model $p$ and approximate posterior $q$ as $J_{\text{ELBO}}:=$
\begin{align}
    \label{eq:elbo_objective_generic}
    \textstyle 
    \mathbb{E}_{q_{\psi}(\theta)} \left[ \sum_{i=1}^{N} \log p_{\eta}(y_i | \theta) \right] - D_{\text{KL}}(q_{\psi}(\theta) \| p_{\eta}(\theta)).
\end{align}
This objective is a function of data $y$, variational parameters $\psi$, and hyperparameters $\eta$.
Maximizing $J_{\text{ELBO}}$ is equivalent to finding $q$ ``closest'' to the true posterior in the sense of Kullback-Leibler (KL) divergence~\citep{blei2017variational}.
As its name suggests, the ELBO is a lower bound on the log of the evidence: $J_{\text{ELBO}}( y_{1:N}, \psi, \eta ) \leq \log \int_{\theta} p_{\eta}( y_{1:N}, \theta) d\theta$. 

\textbf{Target scenario.} We focus on scenarios with two key assumptions.
First, we assume a model as in Eq.~\eqref{eq:joint_pdf_template} where $D \gg N$, where $D$ is the size of parameter $\theta$ and $N$ is size of available training data.
Second, we assume the approximate posterior $q$ is Gaussian with an \textbf{\emph{isotropic}} covariance matrix: $q_{\psi}(\theta) = \mathcal{N}(\theta | \bar{\theta}, \bar{\sigma}_q^2 I_D )$, where $\psi =\{ \bar{\theta}, \sigma_q \}$ with $\bar{\theta} \in \mathbb{R}^D$ and $\sigma_q \in \mathbb{R}_{> 0}$. This last assumption is motivated by scalability.

\begin{figure*}[t!]
    \centering
    \begin{subfigure}{0.3333\textwidth}
        \centering
        \includegraphics[width=\textwidth]{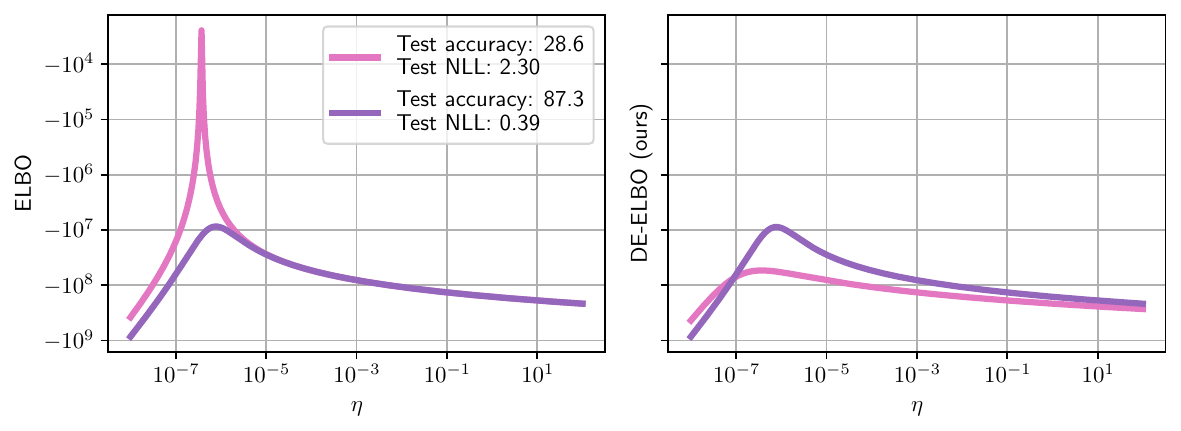}
        \captionsetup{font=scriptsize,labelfont=scriptsize}
        \caption{Model selection as function of $\eta$}
    \end{subfigure}%
    \begin{subfigure}{0.3333\textwidth}
        \centering
        \includegraphics[width=\textwidth]{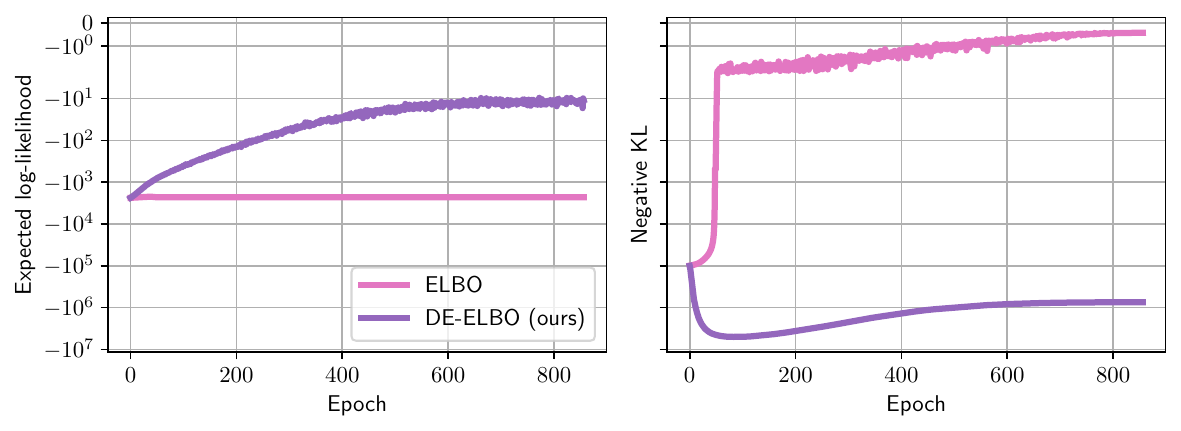}
        \captionsetup{font=scriptsize,labelfont=scriptsize}
        \caption{ELBO terms over training epochs}
    \end{subfigure}%
    \begin{subfigure}{0.3333\textwidth}
        \centering
        \includegraphics[width=\textwidth]{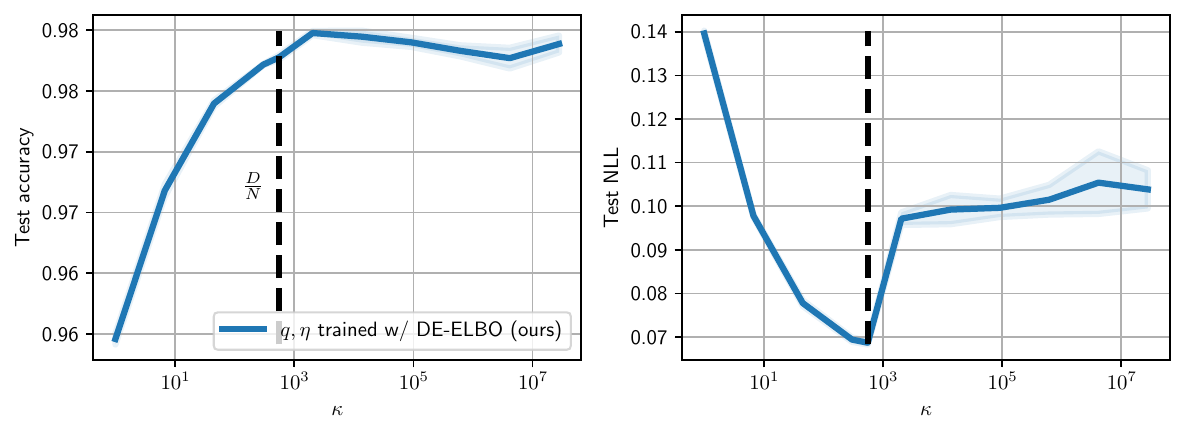}
        \captionsetup{font=scriptsize,labelfont=scriptsize}
        \caption{Performance as function of $\kappa$}
    \end{subfigure}
    \caption{\emph{Panels (a)-(b)}: Comparing approximate posteriors $q$ trained for ELBO (\pink{pink}) and DE-ELBO (ours, \purple{purple}).  Task: ResNet-50 with $D > 1\text{e}6$ trained on CIFAR-10 with $N = 1000$. (a) shows which objective prefers which $q$ across $\eta = \lambda = \tau$. (b) shows which terms in each objective matter most over training steps. \textbf{Takeaway: When $D \gg N$, the ELBO prefers simpler $q$ (\pink{pink}) close to the prior, while our DE-ELBO favors $q$ with higher test accuracy (\purple{purple}).}
    \emph{Panel (c):} Test accuracy and negative log-likelihood (NLL, lower is better) for $q$ trained via our DE-ELBO with various $\kappa$ values. Task: ConvNeXt-Tiny with $D > 1\text{e}6$ trained on CIFAR-10 with $N = 50000$. \textbf{Takeaway: Set $\kappa = \frac{D}{N}$.}
    }
    \label{fig:elbo_comparison}
\end{figure*}

For gradient-based learning, we further assume both $\theta$ and $\eta$ contain only continuous real values. 
If some values are discrete, we can use ELBO (or later, DE-ELBO) objectives, but not gradient-based algorithms.

\textbf{ELBO for $\eta$ selection.}
The fact that ELBO lower bounds the marginal likelihood suggests its utility for selecting hyperparameters $\eta$. 
Work over decades has used the ELBO to select hyperparameters in mixture models~\citep{ueda2002bayesian}, Gaussian processes (GPs;~\citealp{titsias2009variational,damianou2013deep,hensman2015scalable,lalchand2022sparse}), and small neural networks~\citep{kc2021joint}. Yet none of these fit our target scenario, violating either $D \gg N$ or the assumption that $q$ is a Gaussian with simplified covariance for tractability.
\citet{blei2017variational} caution that while the ELBO sometimes works in practice, ``selecting based on a bound is not justified in theory.''
Yet \citet{cherief2019consistency} show that ELBO-based model selection enjoys theoretical guarantees on quality, even under misspecification.

Unfortunately, modern Bayesian deep learning efforts for our target $D \gg N$ scenario remain dominated by time-consuming grid search ~\citep{osawa2019practical,shwartz2022pre,harvey2024transfer,rudner2024finetuning} rather than gradient-of-ELBO learning.

\section{Contributions}
\label{sec:contributions}

We offer two contributions to improve Bayesian model selection in our target scenario. 

\textbf{Contribution 1: When $D \gg N$ and $q$ is a misspecified isotropic Gaussian, the ELBO prefers settings of $\psi,\eta$ that \underline{underfit}}. Thus, gradient-based learning of both $\psi,\eta$ via the ELBO (see Alg.~\ref{alg:deelbo} with $\kappa=1$) often yields notably subpar prediction compared to search methods for $\eta$.

Our evidence for Contribution 1 comes in analytical and experimental forms.
First, in Case Study A (Sec.~\ref{sec:caseA_model_definition}) we analyze a regression model whose ideal (non-isotropic) posterior yields good data fits at small $N$ for arbitrarily large $D$. There, we prove in \hyperlink{lemma1}{Lemma~1} that when $q$ has an isotropic covariance matrix, the ELBO prefers the posterior variance in $\psi$ to match the prior variance as $D$ gets larger but $N$ stays fixed. This too-large variance leads to underfitting. 

\begin{figure*}[!t]
    \begin{center}
        \includegraphics[width=\linewidth]{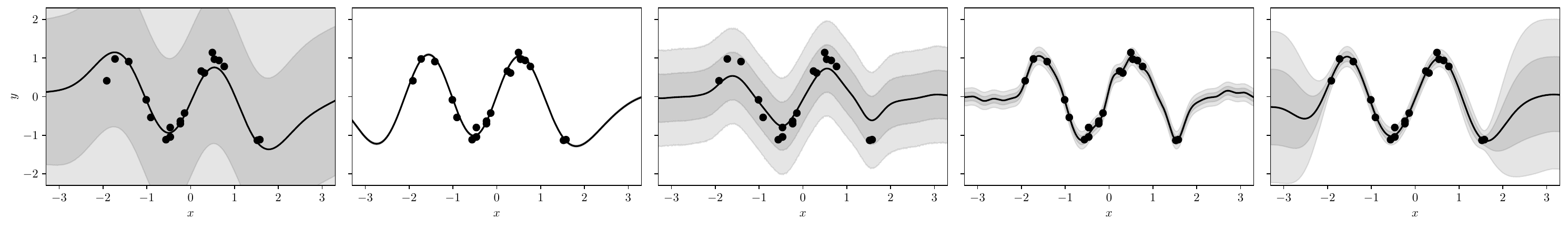}
        \begin{subfigure}{0.1951\linewidth}
            \captionsetup{font=scriptsize,labelfont=scriptsize}
            \subcaption{\makecell{diagEF\\LA-LML}}
        \end{subfigure}
        \begin{subfigure}{0.1951\linewidth}
            \captionsetup{font=scriptsize,labelfont=scriptsize}
            \subcaption{\makecell{diagEF\\LA-CLML}}
        \end{subfigure}
        \begin{subfigure}{0.1951\linewidth}
            \captionsetup{font=scriptsize,labelfont=scriptsize}
            \subcaption{\makecell{isotropic $q$\\ELBO}}
        \end{subfigure}
        \begin{subfigure}{0.1951\linewidth}
            \captionsetup{font=scriptsize,labelfont=scriptsize}
            \subcaption{\makecell{isotropic $q$\\DE-ELBO (ours)}}
        \end{subfigure}
        \begin{subfigure}{0.1951\linewidth}
            \captionsetup{font=scriptsize,labelfont=scriptsize}
            \subcaption{\makecell{$D{\times}D$ covariance\\LML}}
        \end{subfigure}
    \end{center}
    \caption{Predictions using $\psi,\eta$ selected by different objectives for RFF regression.
    We show diagEF LA-LML~\citep{immer2021scalable}, diagEF LA-CLML~\citep{lotfi2022bayesian}, iso ELBO, iso DE-ELBO (ours), and the true posterior for $\eta$ that optimize LML.
    We plot train data $y_{1:N}$ with the mean and two standard deviations of the predictive posterior $p(y_* | y_{1:N})$.
    \textbf{Takeaway: DE-ELBO best approximates the true posterior's mean and variance near data.
    Variance far from data is underestimated.}
    }
    \label{fig:caseA_regression_demo}
\end{figure*}

Second, consider a DNN image classifier for CIFAR-10 data as described later in 
Case Study B. Fig.~\ref{fig:elbo_comparison}~(a) compares two different isotropic $q$ at this task, colored pink and purple. 
The pink $q$ fixes $\psi$ parameters to values that optimize ELBO;  it scores higher ELBO than the purple $q$ across a range of $\eta$.
Yet this model delivers subpar test accuracy of just 28.6\%. 
Throughout later experiments on toy data (see Fig.~\ref{fig:caseA_regression_demo}) and Bayesian transfer learning of image and text classifiers (see Fig.~\ref{fig:computational_time_comparison}), ELBO learning yields subpar accuracy. 

To better understand the reason for underfitting, Fig.~\ref{fig:elbo_comparison}~(b)'s pink lines plot over epochs the two additive terms in Eq.~\eqref{eq:elbo_objective_generic} that define the ELBO. The negative KL term starts out at a large negative value and approaches zero throughout training, suggesting $q$ is approaching the prior. In contrast, the likelihood term makes no visible progress from its modest initial value.
These trace plots motivate a remedy:  upweight the likelihood term.

\textbf{Contribution 2: Emphasizing the data likelihood in the ELBO with upweighting factor $\kappa = \frac{D}{N}$ yields $\psi,\eta$ values that fit data better when $D \gg N$ and $q$ is misspecified.} 
We propose the \emph{data-emphasized ELBO} objective, $J_{\text{DE-ELBO}}:=$
\begin{align}
    \label{eq:de_elbo_objective}
    \textstyle
    \kappa \cdot \mathbb{E}_{q_{\psi}(\theta)} \left[ \sum_{i=1}^{N} \log p_{\eta}(y_i | \theta) \right] {-} D_{\text{KL}}(q_{\psi}(\theta) \| p_{\eta}(\theta))
\end{align}
where we have introduced a scaling factor $\kappa$ on the likelihood term. 
$\kappa=1$ recovers the standard ELBO; instead we recommend larger $\kappa = \frac{D}{N}$ to address the issues raised in Contribution 1. The DE-ELBO is equivalent to a standard ELBO for $\kappa N$ \emph{i.i.d.} data instances, where we happen to observe $\kappa$ copies of dataset $y_{1:N}$.

Gradient-based learning of both $\psi,\eta$ with this objective (see Alg.~\ref{alg:deelbo} with $\kappa=\frac{D}{N}$) fits data better than the standard ELBO or other Bayesian methods. Its prediction quality rivals more expensive search methods in far less time.

Our evidence for Contribution 2 comes in analytical and experimental forms. 
We prove in \hyperlink{lemma2}{Lemma 2} for the Case Study A regression model that DE-ELBO with $\kappa=\frac{D}{N}$ delivers posterior variance that does not collapse to the prior as $D \rightarrow \infty$. 
This means even when $D \gg N$, $\psi, \eta$ can be learned via DE-ELBO to deliver compelling data fit.
In practice, returning to Fig.~\ref{fig:elbo_comparison}, the purple $q$ with $\psi,\eta$ that optimize DE-ELBO with $\kappa=\frac{D}{N}$, has far better test accuracy of 87.3\%. While DE-ELBO prefers this purple solution, standard ELBO clearly does not. 
Finally, Case Study B experiments varying $\kappa$ (see Fig.~\ref{fig:elbo_comparison}~(c), App.~\ref{sec:caseB_varying_kappa}) show that $\kappa=\frac{D}{N}$ has competitive accuracy  across datasets, far better than ELBO ($\kappa=1$).

\section{Related Work}
\label{sec:related_work}

Here we survey other work on modifying ELBOs and learning hyperparameters. Although our objective is mathematically similar to previous work on modified ELBOs~\citep{zhang2018noisy,aitchison2021statistical,mclatchie2025predictive}, our setting of $\kappa = \frac{D}{N}$ and our purpose of enabling gradient-based learning of hyperparameters is distinct.


\textbf{Upweighting data in the ELBO.}
Motivated by different goals than model selection, previous work has also upweighted the likelihood term of the ELBO
via a $\kappa$ multiplier as in Eq.~\eqref{eq:de_elbo_objective}, or equivalently downweighted the KL term.
This line of work \citep{aitchison2021statistical,osawa2019practical,zhang2018noisy,pitas2024fine} refers to reweighting as \emph{tempering}.
They pursue reweighted variational ELBOs to better capture posterior uncertainty, but do not pursue gradient-based learning of $\eta$.
Despite awareness that it is ``favorable to tune regularization'' \citep{zhang2018noisy}, often only a small grid of candidate $\eta$ are searched, as in \citet[Fig. 8]{osawa2019practical} or \citet{zhang2018noisy}, perhaps due to large costs of each separate run.
\citet{aitchison2021statistical} do not tune regularization hyperparameters at all.

\textbf{Downweighting data in the ELBO.}
Other work downweights the likelihood in the ELBO for purposes other than learning $\eta$.
\citet{mandt2016variational}'s variational tempering method downweights data to avoid local optima in mixture models.
Some Bayesian approaches that seek to counter-act model misspecification have effectively downweighted data by raising the likelihood to a power \emph{smaller than one}. This includes the \emph{power likelihood} \citep{antoniano2013bayesian}, \emph{power posterior} \citep{friel2008marginal,miller2019robust}, or ``safe'' Bayesian learning~\citep{grunwald2012safe,grunwald2017inconsistency}. 
Work on $\beta$-variational autoencoders~\citep{higgins2017beta} upweights the KL term of the ELBO, reducing data influence to learn disentangled embeddings.
None of these works learn $\eta$ via gradient ascent as we do.

\textbf{Cold posteriors.}
Other work in Bayesian deep learning has recommended \emph{cold posteriors}~\citep{wenzel2020good, kapoor2022uncertainty}.
This work's objective is mathematically different, as it multiplies the entire log posterior by a scalar temperature, not just the log-likelihood as we do. 
The stated purpose of this temperature is to improve heldout prediction quality of samples from the inferred posterior over $\theta$.
Rather than seeking $q$ from a  variational method, they pursue more expensive sampling-based inference without diagonal or isotropic simplifications.
They do not seek gradient-based learning of hyperparameters $\eta$.

\textbf{Bayesian learning of $\eta$.} \citet{immer2021scalable} offer a leading alternative to the ELBO for learning $\eta$.
They optimize a Laplace approximation of the log marginal likelihood (LA-LML) where the covariance matrix is affordable via a diagonal empirical Fisher (diagEF) approximation.  
We compare to this baseline throughout Sec.~\ref{sec:caseA} and \ref{sec:caseB} below.
It allows gradient-based learning of $\eta$, but can take roughly 2x longer than ELBO or DE-ELBO for a single training run (timings in Tab.~\ref{tab:computational_time_comparison}).

\citet{lotfi2022bayesian} suggest a conditional log marginal likelihood (CLML) objective for selecting hyperparameters. For DNNs, they compare separate runs at distinct $\eta$, but cannot do gradient-based learning.
We compare to their LA-CLML method in Sec.~\ref{sec:caseA} and \ref{sec:caseB}, which reuses \citeauthor{immer2021scalable}'s diagEF Laplace approximation.

\textbf{Gradient methods for $\eta$.}
Some non-ELBO works~\citep{maclaurin2015gradient,lorraine2020optimizing} learn hyperparameters via gradients of validation-set performance metrics. These require large validation sets to perform well, while our DE-ELBO does not use a validation set and may be easier to implement.

\textbf{Smart search for $\eta$.}
Stepping back, if the goal is simply to tune hyperparameters for a point estimation task, 
many smart search strategies have been proposed. Random search \citep{bergstra2012random}, successive halving \citep{karnin2013almost,jamieson2016non}, BO~\citep{snoek2012practical,turner2021bayesian}, or meta-learned BO~\citep{wang2024pre} all have advantages over grid search.
Yet all require separate runs for each candidate $\eta$ and dividing available data into train and validation sets, unlike ELBO-based approaches.


\section{Case Study A: Fourier Regression}
\label{sec:caseA}

The Gaussian process (GP;~\citealp{rasmussen2006gaussian}) is a regression model often paired with a
\emph{radial basis function} (RBF) kernel. This kernel can be defined as $k(x, x') = \sigma_k^2 \exp \left(- \frac{\|x - x'\|_2^2}{2\ell_k^2} \right)$, where $\ell_k, \sigma_k$ are lengthscale and outputscale hyperparameters.
Though GP function complexity can elegantly grow with dataset size $N$, a downside is that fitting scales cubically with $N$.
As a remedy, we consider \emph{random Fourier features} (RFFs;~\citealp{rahimi2007random}), a weight-space model that scales \emph{linearly} in $N$ yet approximates a GP.
The approximation quality increases with a user-controlled size parameter $R$ which sets the overall size: $D = R$.

GPs are notoriously sensitive to hyperparameters $\eta = \{\ell_k, \sigma_k\}$~\citep{rasmussen2006varying}. RFFs are also sensitive, yet tuning $\eta$ well has been largely overlooked. 
\citet{rahimi2007random} fix $\ell_k = 1, \sigma_k = 1$. When \citet{liu2020simple,liu2023simple} used RFF classifiers, they set $\ell_k = 2$ 
and tune $\sigma_k$ 
via search with multiple runs. 

We intend to show how our DE-ELBO yields affordable gradient-based learning of $\ell_k, \sigma_k$ when $D \gg N$, thus improving overall prediction quality.
Our work here could be used as an efficient (\emph{linear in $N$}) GP or as a drop-in way to improve  \citeauthor{liu2020simple}'s distance-aware methods.
Our isotropic $q$s are scalable complements to exact RFF posteriors \citep{potapczynski2021bias}.

\subsection{Model A Definition}
\label{sec:caseA_model_definition}

We train on $N$ pairs $x_i, y_i$ of input vectors $x_i \in \mathbb{R}^H$ and targets $y_i \in \mathbb{R}$.
We first map each $x_i$ to a transformed RFF representation $\phi(x_i) \in \mathbb{R}^{R}$ via
\begin{align}
    \label{eq:phi}
    \textstyle \phi(x_i) = \sigma_k \sqrt{\frac{2}{R}} \cos \left(\frac{1}{\ell_k} A^\top x_i + b\right),
\end{align}
The non-learnable weights $A \in \mathbb{R}^{H \times R}$ and $b \in \mathbb{R}^R$ are randomly drawn \textbf{once} as $A_{h,r} \sim \mathcal{N}(0, 1)$ and $b_{r} \sim \operation{Unif}(0, 2\pi)$ for all $h$ and $r$. $A$ and $b$ remain fixed through all remaining training and prediction.
Past work~\citep{liu2020simple} typically has $R$ in the range $100$ to $10000$.
We set $R = 1024$.

To complete the regression model, we
predict $\hat{y}_i = v^\top \phi(x_i)$ with weights $v \in \mathbb{R}^R$.

\textbf{Contribution: RFFs for arbitrary lengthscale.} 
Our featurization in Eq.~\eqref{eq:phi} generalizes the  construction of RFFs by \citet{rahimi2007random} to any lengthscale $\ell_k > 0$ and outputscale $\sigma_k > 0$.
In App.~\ref{sec:caseA_lengthscale_and_outputscale}, we \emph{\textbf{prove}} that our  construction is a Monte Carlo approximation of the RBF kernel. That is, for any pair of feature vectors $\phi(x_i)^\top \phi(x_j) \approx k(x_i, x_j)$, where $k(\cdot)$ is an RBF kernel whose $\ell_k,\sigma_k$ values match those used to construct $\phi(\cdot)$ in Eq.~\eqref{eq:phi}. The quality of this approximation increases with $R$. Past work has proven that RFF features approximate the RBF kernel only when $\ell_k = 1,\sigma_k = 1$ \citep{rahimi2007random}. To the best of our knowledge, our proof for arbitrary $\ell_k$ is novel.

\textbf{Point estimation view.} To fit the RFF model to  data, empirical risk minimization  seeks weights $v$ that minimize the loss function $L(v) :=$
\begin{align}
    \label{eq:rff_loss_with_l2_penalty}
    \textstyle \sum_{i=1}^N \ell(y_i, v^\top \phi(x_i) ) + \frac{1}{2} || v ||_2 ^2,
\end{align}
where $\ell(\cdot)$ is a loss function (e.g., mean squared error). The L2 penalty on $v$ helps avoid overfitting given many features.
Ultimately, model quality is impacted by hyperparameters $\ell_k > 0, \sigma_k > 0$. Neither can be set effectively by minimizing training loss $L(\cdot)$ alone. We later show how to \emph{learn} $\ell_k,\sigma_k$ using the DE-ELBO.

\textbf{Bayesian view.}
We can define a probabilistic model:
\begin{align}
    \label{eq:rff_joint_pdf_regression_model}
    p(v) &= \mathcal{N}(v | 0_R, I_R), \\ \notag 
    p(y | v) &= \textstyle \prod_{i=1}^N \mathcal{N}(y_i | v^\top \phi(x_i), \sigma_y^2).
\end{align}
This model fits into our general framework from Sec.~\ref{sec:background}: $\theta = \{v\}$, $\eta = \{\sigma_y, \ell_k, \sigma_k \}$, and $D = R$. 
Maximum a-posteriori (MAP) estimation of $v$ recovers the objective in Eq.~\eqref{eq:rff_loss_with_l2_penalty} when we set $\ell(\cdot)$ to $- \log p(y_i | v)$.

\textbf{Ideal posterior.}
For the regression model in Eq.~\eqref{eq:rff_joint_pdf_regression_model}, the true posterior is multivariate Gaussian, with full-rank covariance $\Sigma_{\text{post}} = (I_D + \frac{1}{\sigma_y^2} \Phi^\top \Phi)^{-1}$, where $\Phi \in \mathbb{R}^{N \times D}$ stacks features $\phi(x_i)$ for each of the $N$ train examples. For derivation, see App.~\ref{sec:caseA_regression_model} and \ref{sec:caseA_closed-form_elbo}.

\subsection{Variational Methods for Model A}
\label{sec:caseA_variational}

To apply the general variational recipe described in Sec.~\ref{sec:background} to the model in Eq.~\eqref{eq:rff_joint_pdf_regression_model}, we first select an approximate posterior over parameter $v$.
For simplicity and speed when $D$ is large, we choose a Gaussian with unknown mean and isotropic covariance:
\begin{align}
    q(v) = \mathcal{N}( v | \bar{v}, \bar{\sigma}_q^2 I_D).
\end{align}
Here, the free parameters that define $q$ are $\psi = \{\bar{v}, \bar{\sigma}_q \}$, with $\bar{v} \in \mathbb{R}^D$ and $\bar{\sigma}_q \in \mathbb{R}_{> 0}$. 

\subsection{Theoretical Analysis of Model A}

First, in Lemma 1 below we show that for over-parameterized models the assumption of isotropic $q$ leads to undesirable underfitting when optimizing the ELBO. Next, we show the DE-ELBO avoids this issue, particularly by setting $\kappa = \frac{D}{N}$.

\hypertarget{lemma1}{\textbf{Lemma 1.}} Assuming isotropic $q$, as $D \rightarrow \infty$ the optimal approximate posterior variance $\bar{\sigma}_q^2$ for the ELBO ($\kappa = 1$) exactly matches the prior variance of $1$.

\textbf{Proof.} Set $\nabla_{\bar{\sigma}_q^2} J_{\text{ELBO}} = 0$, solve for $\bar{\sigma}_q^2$, then take the limit:
\begin{align}
    \bar{\sigma}_q^{2*} = \frac{D}{\frac{1}{\sigma_y^2}\operation{tr}(\Phi\Phi^\top) + D},~\lim_{D\rightarrow\infty} \bar{\sigma}_q^{2*} = 1.
    \notag
\end{align}
By construction, we have $\Phi \Phi^\top \approx K$, where $K$ is the $N$-by-$N$ matrix of kernel evaluations on the train set. For the chosen RBF kernel, $\operation{tr}(K) = \sigma_k^2 N$. 
$\blacksquare$

From Lemma 1, we conclude that as parameter size $D$ increases but $N$ remains fixed, the variance of the approximate posterior $q$ approaches a value of 1, matching the isotropic prior in Eq.~\eqref{eq:rff_joint_pdf_regression_model}.
As a result, RFF regressors trained to maximize ELBO can underfit, preferring higher posterior variance than may be needed (see Fig.~\ref{fig:caseA_regression_demo}~(c)).
Similar underfitting has been shown for BNNs~\citep{coker2022wide}.

\hypertarget{lemma2}{\textbf{Lemma 2.}} In the same setting as Lemma 1, as $D \rightarrow \infty$ the optimal approximate posterior variance for the DE-ELBO ($\kappa = \frac{D}{N}$) will be smaller than the prior variance.

\textbf{Proof.} Set $\nabla_{\bar{\sigma}_q^2} J_{\text{DE-ELBO}} = 0$, solve for $\bar{\sigma}_q^2$, then take the limit:
\begin{align}
    \bar{\sigma}_q^{2*} = \frac{D}{\frac{D}{N} \frac{1}{\sigma_y^2}\operation{tr}(\Phi\Phi^\top) + D},~\lim_{D\rightarrow\infty} \bar{\sigma}_q^{2*} = \frac{1}{\frac{\sigma_k^2}{\sigma_y^2} + 1} < 1.~\blacksquare
    \notag
\end{align}
The DE-ELBO thus helps the misspecified posterior retain dependence on the data even when $D \gg N$, avoiding collapse to the prior. The particular choice of $\kappa = \frac{D}{N}$ is key to this result via elegant cancellation in the denominator.
When hyperparameters $\sigma_y, \ell_k, \sigma_k$ are \emph{learnable}, DE-ELBO can
produce high-quality fits to data even for large $D$, as in Fig.~\ref{fig:caseA_regression_demo}~(d).

\subsection{Experiments for Model A}

We compare different methods for training $\psi,\eta$ for RFF models with $R = D = 1024$ on an $N = 20$ univariate regression dataset $y = \sin(3x) + \varepsilon$ where $\varepsilon \sim \mathcal{N}(0, 0.01)$.
The goal here is to illustrate sensitivity to hyperparameters $\eta = \{\sigma_y, \ell_k,\sigma_k\}$ and the effective learning of $q, \eta$ enabled by our approach.

\textbf{Baselines.}
We fit isotropic (iso) posteriors $q$ and hyperparameters $\eta$ via the standard ELBO and our proposed DE-ELBO, via Alg.~\ref{alg:deelbo}. 
We compare to diagonal (diag) covariance versions of LA-LML \citep{immer2021scalable} and LA-CLML \citep{lotfi2022bayesian}.

\textbf{Common training plan.}
For all methods on this toy dataset, we perform gradient ascent using all training data in every step without any data augmentation.
We train for a specified number of iterations and verify convergence by inspection.
Each run of our method and the baselines depends on the adequate selection of learning rate.
All runs search over 4 candidate values, selecting the best using a method-appropriate objective (e.g., train set DE-ELBO for iso DE-ELBO).
See pseudocode in App.~\ref{sec:pseudocode} for details.

\textbf{Learning hyperparameters $\eta = \{ \sigma_y, \ell_k, \sigma_k \}$.}
For ELBO and DE-ELBO, at each epoch we update variational parameters $\psi$ and hyperparameters $\sigma_y, \ell_k, \sigma_k$ via a gradient step.
Gradients of each ELBO-based objective $J$ are easily computed with respect to $\sigma_y, \ell_k, \sigma_k$ via automatic differentiation.
We use the softplus reparameterization to handle positivity constraints.

\subsection{Results for Model A}
Fig.~\ref{fig:caseA_regression_demo} visualizes the posterior predictive distribution for each method on the $N=20$ toy regression task.
The iso ELBO underfits with high variance, validating our Lemma 1.
Our iso DE-ELBO best approximates the true predictive posterior's mean and variance near data, outperforming the other baselines even though they have flexibility to use a diagonal (not isotropic) covariance.
The iso DE-ELBO does \emph{underestimate} the variance far from data.
This is expected since we chose an isotropic $q$ for tractability not high fidelity to the posterior.
App.~\ref{app:caseA} has more RFF results on \emph{classification}.
\section{Case Study B: Transfer Learning}
\label{sec:caseB}

\subsection{Model B Definition}

For Case Study B, we explore transfer learning of image and text classifiers using informative priors~\citep{li2018explicit,shwartz2022pre} in our target scenario.
Each neural network has two parts. First, a backbone encoder $f(\cdot)$ with weights $w \in \mathbb{R}^F$ maps input vector $x_i$ to a representation vector $z_i \in \mathbb{R}^H$.
For transfer learning, we assume the backbone weights $w$ are high-dimensional ($F$ is very large) and that $w$ has been pre-trained to a high-quality initial value $\mu$ on a source task. 
Second, a linear-boundary classifier  with weights $V \in \mathbb{R}^{C \times H}$ leads to probabilities over $C$ possible classes. We seek values of $w$ and $V$ that classify well on a provided \emph{target task} dataset of $N$ pairs $x_i, y_i$ of features $x_i$ and corresponding class labels $y_i \in \{1, 2, \ldots C\}$.

\textbf{Deep learning view.}
Typical DNN approaches to transfer learning (e.g., baselines in \citet{li2018explicit}) would pursue empirical risk minimization with regularization, training to minimize the loss $L(w, V) :=$
\begin{align}
  \label{eq:loss_with_l2_penalty}
  \textstyle \sum_{i=1}^N \ell(y_i, V f_w( x_i) ) + \frac{\alpha}{2} || w ||_2 ^2 + \frac{\beta}{2} || \operation{vec}(V) ||_2 ^2 
\end{align}
where $\ell(\cdot)$ is a cross-entropy loss indicating agreement with the true label $y_i$, while the L2-penalty on weights $w,V$ favors magnitudes closer to zero, often referred to as ``weight decay''.
Hyperparameters $\alpha \geq 0$, $\beta \geq 0$ control the strength of the L2 penalty.

\textbf{Bayesian view.} For this problem, we can define a joint $p(y_{1:N}, w, V)$ decomposed as in Eq.~\eqref{eq:joint_pdf_template}, where
\begin{align}
    \label{eq:caseB_joint_pdf}
    p(w) &= \mathcal{N}( w | \mu_p, \lambda \Sigma_p), \\
    p(V) &= \mathcal{N}( \operation{vec}(V) | 0_{HC}, \tau I_{HC}), \notag \\
    p(y_i | w, V) &= \operation{Cat}( y_i | \textsc{sm}( V f_w(x_i) ) ). \notag
\end{align}
Here, $\lambda > 0, \tau > 0$ are hyperparameters, $\mu_p, \Sigma_p$ represent \emph{a priori} knowledge of the mean and covariance of the backbone weights $w$, $\textsc{sm}(\cdot)$ is the softmax function, and $\operation{Cat}(\cdot)$ is the categorical probability mass function.
Pursuing MAP estimation for $w$ and $V$ recovers the objective in Eq.~\eqref{eq:loss_with_l2_penalty} when we set
$\alpha = \frac{1}{\lambda}, \beta = \frac{1}{\tau}, \mu_p = 0_F, \Sigma_p = I_F$, and $\ell(\cdot)$ to $- \log p(y_i | w, V)$. In terms of our general framework, we have $\theta = \{w, V\}$, $\eta = \{\lambda, \tau\}$, and $D = F + HC$.

\textbf{Need for validation set and grid search.}
Selecting $\alpha,\beta$ (or equivalently $\lambda,\tau$) to directly minimize Eq.~\eqref{eq:loss_with_l2_penalty} on the training set alone is not a coherent way to guard against overfitting.
Regardless of data or weights, we would select $\alpha^* = 0,\beta^* = 0$ to minimize $L(\cdot)$ as a function of $\alpha,\beta$ and thus enforce no penalty on weight magnitudes at all.
We see similar results with $\kappa \gg \frac{D}{N}$ (see App.~\ref{sec:caseB_varying_kappa}).
Carving out a validation set for selecting these hyperparameters is thus critical to avoid overfitting when point estimating $w,V$.

\setlength{\tabcolsep}{2pt}
\begin{wraptable}[7]{R}{115.6362pt}
    \centering
    \caption{Possible priors.}
    \label{tab:transfer_learning_methods}
    \small
    \begin{tabular}{lcc}
        \hline
        \bfseries Method & \bfseries $p(w)$ & Init. \\
        \hline
        L2-zero & $\mathcal{N}(0_F, \lambda I_F)$ & $\mu$ \\
        L2-SP & $\mathcal{N}(\mu, \lambda I_F)$ & $\mu$ \\
        PTYL & $\mathcal{N}(\mu, \lambda \Sigma)$ & $\mu$ \\
        \hline
    \end{tabular}
\end{wraptable}
\setlength{\tabcolsep}{6pt}

\textbf{Backbone priors.}
Several recent transfer learning methods correspond to specific values of the mean and covariance $\mu_p, \Sigma_p$ of the backbone prior $p(w)$. Let $\mu$ represent pre-trained backbone weights from the source task.
Setting $\mu_p = 0_F, \Sigma_p = I_F$ recovers a conventional approach we call L2-zero, where regularization pushes backbone weights to zero. The pre-trained $\mu$ only informs the initial value of $w$ before stochastic gradient descent (SGD).
~Instead, setting $\mu_p = \mu, \Sigma_p = I_F$ recovers \emph{L2 starting point} (L2-SP) regularization~\citep{chelba2006adaptation,li2018explicit}. Further setting $\Sigma_p$ to the covariance matrix of a Gaussian approximation of the posterior over backbones from the source task recovers the ``Pre-Train Your Loss'' (PTYL) method~\citep{shwartz2022pre}.

\textbf{Need to specify a search space.}
Selecting $\alpha,\beta$ (or equivalently $\lambda,\tau$) via grid search requires  specifying a grid of candidates spanning a finite range.
For PTYL, the optimal search space for these hyperparameters is still unclear.
For the same prior and the same datasets, the search space has varied between works: PTYL's creators recommended large values from 1e0 to 1e10~\citep{shwartz2022pre}.
Later works search smaller values (1e-5 to 1e-3)~\citep{rudner2024finetuning}.
Our DE-ELBO with per-epoch learning of $\eta$ avoids the need to set any  predefined ranges.

\subsection{Variational Methods for Model B}
To apply the general variational recipe described in Sec.~\ref{sec:background} to Model B, we first select an approximate posterior over parameters $w, V$. For tractability and speed, we choose a factorized Gaussian with unknown means and isotropic covariance controlled by scalar $\bar{\sigma}_q > 0$:
\begin{align}
    q(w, V) &= q(w)q(V) \\ \notag
    q(w) &= \mathcal{N}( w | \bar{w}, \bar{\sigma}_q^2 I_F), \\ \notag 
    q(V) &= \mathcal{N}( \operation{vec}(V) | \operation{vec}(\bar{V}), \bar{\sigma}_q^2 I_{H C}).
\end{align}
Here, the free parameters for $q$ are $\psi = \{\bar{w}, \bar{V}, \bar{\sigma}_q \}$.

\textbf{ELBO-based training.}
Given a training set, we optimize $\psi, \eta$ to maximize the ELBO or our DE-ELBO via Alg.~\ref{alg:deelbo}. We evaluate the KL term inside each objective in closed-form, as both prior and $q$ are Gaussian. To evaluate the expected log-likelihood term, we use Monte Carlo averaging of $S$ samples from $q$ \citep{xu2019variance,mohamed2020monte}. We find that just one sample ($S=1$) per training step is sufficient and fast. We use the \emph{reparameterization trick}~\citep{blundell2015weight} to obtain gradient estimates for this likelihood term.

\textbf{Learning hyperparameters $\eta = \{ \lambda, \tau \}$.}
While we could use gradients to update the prior variances $\lambda, \tau$, inspecting the closed-form of the KL term in the objective reveals an analytical update guaranteed to deliver the best possible value of $\lambda$ (in terms of ELBO or DE-ELBO) given the current value of $\psi = \{ \bar{w}, \bar{V}, \bar{\sigma}_q \}$.
Setting $\nabla_\lambda J = 0$ and solving for $\lambda$, we get 
\begin{align}
    \lambda^* = \textstyle \frac{1}{F} \left[ \bar{\sigma}_q^2 \operation{tr}(\Sigma_p^{-1}) + (\mu_p-\bar{w})^\top \Sigma_p^{-1} (\mu_p-\bar{w}) \right]
    \label{eq:lambda_update}
\end{align}
Similar updates can be derived for $\tau$ (see App.~\ref{sec:caseB_learning_lambda_tau}).
We use these updates in Line 6 of Alg.~\ref{alg:deelbo}.

\setcounter{figure}{2}
\begin{figure*}[htbp!]
    \includegraphics[width=\linewidth]{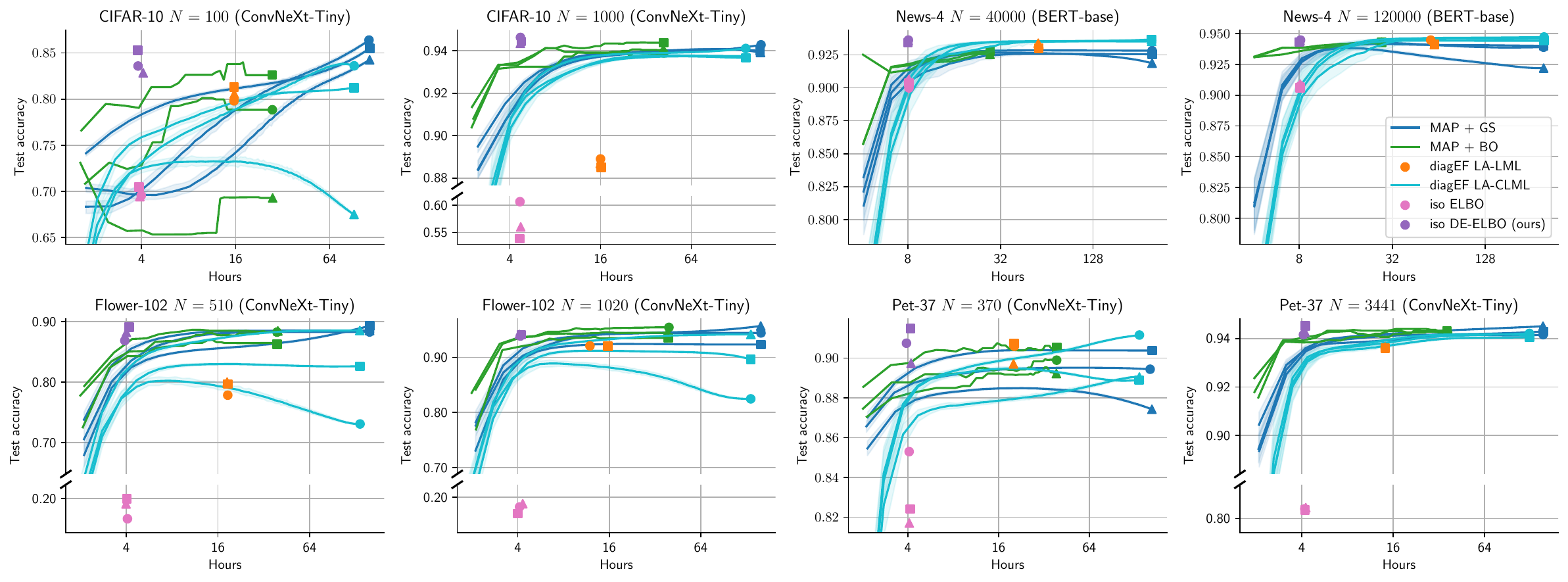}
    \caption{
Test accuracy over time for L2-SP transfer learning of image and text classifiers. 
We run each method on 3 separate train sets of size $N$ (3 marker styles).
Each panel shows a distinct task: ConvNeXt-Tiny fine-tuned on CIFAR-10, Flower-102, and Pet-37; BERT-base fine-tuned on News-4.
We compare MAP + GS, MAP + BO~\citep{hvarfner2024vanilla}, diagEF LA-LML~\citep{immer2021scalable}, diagEF LA-CLML~\citep{lotfi2022bayesian},  iso ELBO, and iso DE-ELBO.
\textbf{Takeaway: After just a few hours, iso DE-ELBO reaches as good or better performance at small data sizes and similar performance at large sizes, even when other methods are given many additional hours.}
Further results in App.~\ref{app:caseB} examine ConvNeXt-Tiny (Fig.~\ref{fig:convnext_tiny_computational_time_comparison}), ViT-B/16 (Fig.~\ref{fig:vit_b_16_computational_time_comparison}), ResNet-50 (Fig.~\ref{fig:resnet_50_computational_time_comparison}), and BERT-base (Fig.~\ref{fig:bert_base_computational_time_comparison}).
    }
\label{fig:computational_time_comparison}
\end{figure*}

\subsection{Experiments for Model B}

\textbf{Image experiments.}
We fine-tune 3 architectures: ResNet-50~\citep{he2016deep}, ViT-B/16~\citep{dosovitskiy2021image}, and ConvNeXt-Tiny~\citep{liu2022convnet}, on 3 datasets with 8 different dataset sizes: CIFAR-10~\citep{krizhevsky2010cifar} with $N \in \{100, 1000, 10000, 50000\}$; Flower-102~\citep{nilsback2008automated} with $N \in \{510, 1020\}$; and Pet-37~\citep{parkhi2012cats} with $N \in \{370, 3441\}$. 
All 3 architectures were pre-trained on ImageNet.

\textbf{Text experiments.}
We fine-tune BERT-base~\citep{devlin2019bert} on News-4~\citep{zhang2015character} with $N \in \{400, 4000, 40000, 120000\}$.
BERT-base was pre-trained on BooksCorpus and English Wikipedia.

\textbf{Baselines.}
We compare our iso DE-ELBO to its natural ablation the iso ELBO, the gradient-based diagEF LA-LML \citep{immer2021scalable}, and the grid search-based diag EF LA-CLML~\citep{lotfi2022bayesian}.

We also compare to MAP + grid search (GS), which does MAP point estimation of $w, V$ via separate SGD runs for each candidate $\lambda, \tau$ value in a fixed grid (see App.~\ref{sec:caseB_implementation_details}), selecting the best according to the validation set likelihood.
This GS baseline represents cutting-edge work in transfer learning~\citep{shwartz2022pre,harvey2024transfer}.
We further compare to a smarter search method: MAP + Bayesian optimization (BO;~\citealp{hvarfner2024vanilla}).

\textbf{Common training plan.}
For each dataset size, we draw 3 separate random training sets of size $N$ from the full training set, stratifying by class to ensure balanced class frequencies. We run each method on all 3 sets. 

Results for diagEF LA-LML, iso ELBO, and iso DE-ELBO do not use a validation set.
For all other methods, we hold out $\frac{1}{5}$ of the training set for validation, stratifying by class to ensure balanced class frequencies. 
After selecting the best hyperparameters, we retrain the model using the selected hyperparameters on the combined training and validation set.

For all methods, we perform minibatch SGD with a Nesterov momentum parameter of 0.9.
For image experiments, we use a batch size of 128 with light data augmentation (random crops and horizontal flips) and train for 6000 steps using a cosine annealing learning rate~\citep{loshchilov2016sgdr}.
For text experiments, we use a batch size of 32 without any data augmentation and train for 12000 steps using a cosine annealing learning rate.
Each run of our method and the baselines depends on the adequate selection of learning rate. All runs search over 4 candidate values and select the best according to a method-appropriate objective.
See pseudocode in App.~\ref{sec:pseudocode} for details on learning rate selection for the ELBO and our DE-ELBO.

\textbf{Estimating ELBO and accuracy.}
After training, we estimate the expected log-likelihood term of the ELBO or our DE-ELBO for model selection by averaging over 10 samples from $q$.
To compute classifier accuracy given $q$, we find that just plugging in the means $\bar{w}, \bar{V}$ to make predictions gives similar accuracy to averaging over 10 posterior samples without the added runtime cost.

\setlength{\tabcolsep}{2pt}
\begin{table*}[t!]
  \caption{Timing for L2-SP transfer learning methods. Task: ResNet-50 fine-tuned on CIFAR-10 with $N = 50000$. We compare MAP + GS, diagEF LA-LML \citep{immer2021scalable}, and iso DE-ELBO (ours). See App.~\ref{sec:caseB_implementation_details} for search details. 
  \textbf{Takeaway: Each iso DE-ELBO run is 2x faster than diagEF LA-LML and learns $\lambda,\tau$, avoiding the extreme costs of grid search.}
  Hardware: 4 Intel Xeon 6226R CPUs (2.90 GHz) and 1 NVIDIA A100 GPU (40 GB).
  }
  \label{tab:computational_time_comparison}
  \centering
  \scriptsize
  \begin{tabular}{llc|cc|cc}
    & \multicolumn{2}{c}{\textbf{Size of grid search (GS) space}} & \multicolumn{2}{c}{L2-SP} & \multicolumn{2}{c}{PTYL} \\    
    \textbf{Method} & $\lambda, \tau$ & Learning rate & \textbf{Avg. SGD run} & \textbf{Total GS time} & \textbf{Avg. SGD run} & \textbf{Total GS time} \\
    \hline
    MAP + GS & 36 (L2-SP) / 60 (PTYL) & 4 & 37 min. & \phantom{0}88.5 hr. & 37 min. & 148.7 hr. \\        
    diagEF LA-LML &  Learned via gradients & 4 & 70 min. & \phantom{00}4.7 hr. & \multicolumn{2}{c}{\emph{PTYL prior covariance not supported}} \\ 
    iso DE-ELBO (ours) & Learned via Eq.~\eqref{eq:lambda_update} & 4 & 33 min. & \phantom{00}2.2 hr. & 36 min. & \phantom{00}2.4 hr. \\
  \end{tabular}
\end{table*}
\setlength{\tabcolsep}{6pt}

\subsection{Results for Model B}

Across text and image datasets, several backbones, and several priors, our findings are:

\textbf{The runtime of iso DE-ELBO is affordable, avoiding the extreme time costs of grid search.}
In Tab.~\ref{tab:computational_time_comparison}, an individual SGD run of our iso DE-ELBO has comparable cost to one SGD run of standard MAP estimation. However, the cumulative cost of selecting $\eta$ via grid search is far higher than our approach: 
L2-SP's search from \citet{li2018explicit} takes over 88 hours while PTYL's search \citep{shwartz2022pre} takes over 148 hours.
Our iso DE-ELBO delivers in under 3 hours in both cases by learning $\eta$ via gradients.

\textbf{The accuracy of iso DE-ELBO is competitive with or better than recent Bayesian methods.}
Fig.~\ref{fig:computational_time_comparison} shows test set accuracy over time for L2-SP transfer learning methods on select datasets and architectures for both image and text tasks. 
Our iso DE-ELBO consistently performs competitive with or better than recent Bayesian methods like \citet{immer2021scalable} or \citet{lotfi2022bayesian}, even though they use diagonal (not isotropic) posteriors.
\citet{lotfi2022bayesian}'s DiagEF LA-CLML performs similar to MAP + GS.
For final test accuracy, negative log-likelihood (NLL), and expected calibration error (ECE) results, see App.~\ref{sec:caseB_results}.

\section{Discussion and Conclusion}

We have proposed a practical approach to Bayesian model selection. Our modified ELBO objective enables per-epoch updates to hyperparameters on the full training set.
Our solution is intended for a specific target scenario: where $D \gg N$ and $q$ is simpler than the true posterior for tractability. 
In such scenarios, we provide analytical and experimental evidence for why the standard ELBO yields poor fits to the data, while our \emph{data-emphasized ELBO} delivers better fits. We hope Bayesian deep learning researchers appreciate our approach's favorable comparisons to \citet{immer2021scalable} and \citet{lotfi2022bayesian}. We hope DNN practitioners appreciate the method's reliability at delivering accurate classifiers quickly while avoiding the variability of validation sets when $N$ is small.

\textbf{Limitations.} The per-epoch updates to $\eta$ in our DE-ELBO approach only work for continuous hyperparameters that explicitly appear in the prior or likelihood of a probabilistic model. Separate runs for each candidate would still be needed for discrete hyperparameters or optimization hyperparameters like learning rates, because gradients aren't available. DE-ELBO can still select among candidate runs for such values, as we do for learning rate in Alg.~\ref{alg:deelbo}. 

Though we focus on Gaussian $q$ with isotropic covariance, we conjecture the DE-ELBO would work reasonably with diagonal covariance.
Further work is needed to consider low-rank or full-rank covariances, or non-Gaussian $q$.
The closed-form prior variance updates are only possible because of the closed-form KL term.

More rigorous theoretical understanding of the DE-ELBO is needed, including understanding of weaknesses like the underestimated variance far from data in Fig.~\ref{fig:caseA_regression_demo}~(d), a common issue in variational inference \citep{wilson2020bayesian}. Our recommendation to set $\kappa = \frac{D}{N}$, though supported by Lemma 2 and experiments like Fig.~\ref{fig:convnext_tiny_varying_kappa_acc} and \ref{fig:convnext_tiny_varying_kappa_nll}, could use further support to understand its general applicability to other models or scenarios beyond those studied here.
Especially in tasks that benefit from heavy data augmentation, adjusting the value of $\kappa$ to account for this could be fruitful.

\textbf{Outlook.} 
The DE-ELBO avoids the need for any validation set and expensive separate runs.
It can offer hours of saved time to practitioners, which could be used to further improve models. 
For example, on Pet-37 we found that L2-zero using an initialization from supervised pre-training results in an accuracy gain of 32.4 percentage points over self-supervised pre-training.
Beyond saving valuable time, we hope our work sparks interest in theoretical understanding of modified ELBOs for improved model selection.

\section*{Acknowledgments}
Authors EH and MCH gratefully acknowledge support in part from the Alzheimer’s Drug Discovery Foundation and the National Institutes of Health (grant \# R01NS134859). MCH is also supported in part by the U.S. National Science Foundation (NSF) via grant IIS \# 2338962. We are thankful for computing infrastructure support provided by Research Technology Services at Tufts University, with hardware funded in part by NSF award OAC CC* \# 2018149. We would like to thank Tim G. J. Rudner for helpful comments on an earlier draft of this paper.

\bibliographystyle{plainnat}
\bibliography{main}

\section*{Checklist}



\begin{enumerate}

  \item For all models and algorithms presented, check if you include:
  \begin{enumerate}
    \item A clear description of the mathematical setting, assumptions, algorithm, and/or model. Yes.
    \item An analysis of the properties and complexity (time, space, sample size) of any algorithm. Yes, see Tab.~\ref{tab:computational_time_comparison}.
    \item (Optional) Anonymized source code, with specification of all dependencies, including external libraries. Yes, see URL on page 1.
  \end{enumerate}

  \item For any theoretical claim, check if you include:
  \begin{enumerate}
    \item Statements of the full set of assumptions of all theoretical results. Yes, see Sec.~\ref{sec:caseA_variational}.
    \item Complete proofs of all theoretical results. Yes, see App.~\ref{sec:caseA_closed-form_elbo_isotropic_q} and \ref{sec:caseA_closed-form_deelbo_isotropic_q}.
    \item Clear explanations of any assumptions. Yes, see Sec.~\ref{sec:caseA_variational}.
  \end{enumerate}

  \item For all figures and tables that present empirical results, check if you include:
  \begin{enumerate}
    \item The code, data, and instructions needed to reproduce the main experimental results (either in the supplemental material or as a URL). Yes, see URL on page 1.
    \item All the training details (e.g., data splits, hyperparameters, how they were chosen). Yes, see App.~\ref{sec:caseB_implementation_details}.
    \item A clear definition of the specific measure or statistics and error bars (e.g., with respect to the random seed after running experiments multiple times). Yes, see captions of Tab.~\ref{tab:image_acc}, \ref{tab:image_nll}, \ref{tab:text_acc}, and \ref{tab:text_nll}.
    \item A description of the computing infrastructure used. (e.g., type of GPUs, internal cluster, or cloud provider). Yes, see caption of Tab.~\ref{tab:computational_time_comparison}.
  \end{enumerate}

  \item If you are using existing assets (e.g., code, data, models) or curating/releasing new assets, check if you include:
  \begin{enumerate}
    \item Citations of the creator If your work uses existing assets. Yes, we cite the creators of used assets.
    \item The license information of the assets, if applicable. Yes, see license at URL on page 1.
    \item New assets either in the supplemental material or as a URL, if applicable. Yes, see URL on page 1.
    \item Information about consent from data providers/curators. Yes, datasets are publicly available.
    \item Discussion of sensible content if applicable, e.g., personally identifiable information or offensive content. Not Applicable.
  \end{enumerate}

  \item If you used crowdsourcing or conducted research with human subjects, check if you include:
  \begin{enumerate}
    \item The full text of instructions given to participants and screenshots. Not Applicable.
    \item Descriptions of potential participant risks, with links to Institutional Review Board (IRB) approvals if applicable. Not Applicable.
    \item The estimated hourly wage paid to participants and the total amount spent on participant compensation. Not Applicable.
  \end{enumerate}

\end{enumerate}

\newpage
\appendix
\onecolumn

\counterwithin{table}{section}
\setcounter{table}{0}
\counterwithin{figure}{section}
\setcounter{figure}{0}
\counterwithin{algorithm}{section}
\setcounter{algorithm}{0}

\section*{Appendix}

\addtocounter{section}{2}
\section{Pseudocode}
\label{sec:pseudocode}
Below in Alg.~\ref{alg:deelbo}, we give a practical procedure for estimating  an approximate posterior $q$ and key hyperparameters given a dataset of $N$ observations using our DE-ELBO objective.

This algorithm can handle two kinds of hyperparameters:
\begin{itemize}[noitemsep,leftmargin=*,topsep=0pt]
    \item model hyperparameters $\eta$ that impact the prior or likelihood, via gradient/closed-form updates each epoch,
    \item and other hyperparameters like learning rates that impact optimization quality, via grid search.
\end{itemize}
For the former, we can do gradient-based learning of $\eta$ (see bottom case of Line 6 in code below), or sometimes closed-form updates when setting the analytical gradient to zero and solving is feasible (see Case Study B for an example, especially App.~\ref{sec:caseB_learning_lambda_tau}). For the latter, DE-ELBO can be used to select learning rate, but the objective is not an explicit function of learning rate and so gradient-based learning is not possible. Instead, the outer loop in the code below compares ultimate objective function values at a fixed grid of lr values.

For simplicity, we assume in the pseudocode that the dataset $N$ is small enough that stochastic minibatches are not necessary. 
It is straightforward to extend this algorithm to minibatches (see subsection below). 

The algorithm uses automatic differentiation (AD) for computing all gradients, as well as the reparameterization trick for estimating gradients of the expected log-likelihood term of the ELBO. We use the closed-form KL between the two multivariate Normal distributions (which presumes the prior and $q$ are Normal).

\newcommand{\lr}{\mathsf{lr}}
\newcommand{\lrSet}{C_{\mathsf{lr}}}

\begin{algorithm}[!h]
\caption{Gradient Ascent to Estimate $q$ and Hyperparameters with the Data-Emphasized ELBO}
\label{alg:deelbo}
\textbf{Input}: 
\begin{itemize}[align=left,style=nextline,leftmargin=*,labelsep=3\parindent,font=\normalfont,topsep=0pt,itemsep=-1ex,partopsep=1ex,parsep=1ex]
\item ~training set $\{y_i\}_{i=1}^N$
\item ~likelihood $p_\eta(y_i|\theta)$
\item ~prior $p_\eta(\theta)$
\item ~initial value $\psi_0$ for parameters $\psi = \{ \bar{\theta}, \bar{\sigma}_q \}$ that define the approximate posterior $q_\psi(\theta) = \mathcal{N}( \theta | \bar{\theta}, \bar{\sigma}_q^2 I_D )$
\item ~initial value $\eta_0$ for hyperparameters $\eta$
\item ~data-emphasis factor $\kappa$. Set $\kappa = \frac{D}{N}$ for our recommended DE-ELBO. Instead, $\kappa=1$ recovers standard ELBO.
\item ~set of candidate learning rates $\lrSet$
\end{itemize} 
\textbf{Output}:
\begin{itemize}[noitemsep,leftmargin=*,topsep=0pt]
\item ~estimated parameters $\psi$ for the approximate posterior $q$
\item ~estimated hyperparameters $\eta$
\end{itemize}
\textbf{Procedure}:
\linespread{1.1}\selectfont 
\begin{algorithmic}[1] 
    \STATE \textbf{for} each $\lr~\textnormal{in}~\lrSet$:
    \STATE \hspace{3mm} $\psi \gets \psi_0, \eta \gets \eta_0$
    \STATE \hspace{3mm} \textbf{for} each epoch until converged:
    \STATE \hspace{6mm} $\varepsilon \sim \mathcal{N}(0_D, I_D)$ ~~ \emph{We do just 1 MC sample per train step; could do more if affordable.}
    \STATE \hspace{6mm} $\psi \leftarrow \psi + \lr \left( 
    \kappa \cdot \sum_{i=1}^{N} 
        \underbrace{\nabla_{\psi} \log p_{\eta}(y_i | \theta = \bar{\theta} + \bar{\sigma}_q \epsilon )}_{\tiny \textnormal{Reparameterization trick and AD}}
    ~-~ \underbrace{\nabla_{\psi} D_{\text{KL}} (q_\psi(\theta) \| p_{\eta}(\theta) )}_{\tiny \textnormal{AD on KL of two Normals}}
    \right)$
    \STATE \hspace{6mm} $\eta \leftarrow \begin{cases}
            \eta^*(\psi, \{ y_i \}_{i=1}^N ) & \emph{if closed-form update exists}
        \\
        \eta + \lr \left( \kappa \cdot \sum_{i=1}^{N} \nabla_{\eta} \log p_{\eta}(y_i | \theta=\bar{\theta} + \bar{\sigma}_q \epsilon) - \nabla_{\eta} D_{\text{KL}}(q_\psi(\theta) \| p_{\eta}(\theta) ) \right)
        & \emph{otherwise}
    \end{cases}$ 
    \STATE \hspace{3mm} Calculate $J_{\textsc{DE-ELBO}}$ for final values of $\psi,\eta$ on training set, using 10 MC samples for log-likelihood term
    \STATE Return values $\psi,\eta$ corresponding to $\lr$ that maximized $J_{\textsc{DE-ELBO}}$
\end{algorithmic}
\end{algorithm}

Choosing the number of samples to use at each training step and for the final learning rate selection decision is up to the user. We recommend using the largest values that can complete training in affordable time limits. We used 1 and 10 samples, respectively, throughout our experiments.

\subsection{Handling Positivity Constraints}
The gradient updates as written in Alg.~\ref{alg:deelbo} do not respect the constrained domains of some parameters. For the chosen isotropic Gaussian family for $q$, within $\psi$ there is a scalar variance that must be positive: $\bar{\sigma}_q > 0$. 
Similarly, within $\eta$ there are often continuous hyperparameters with constrained domains (e.g., lengthscales and outputscales must be positive in Case Study A, variances must be positive in Case Study B).

To handle positivity constraints, we reparameterize in terms of unconstrained parameters via a one-to-one mapping.
For example: $\bar{\sigma}_q = \operation{softplus}( \bar{u}_q )$, where $\bar{u}_q \in \mathbb{R}$ and the softplus function is defined as in the PyTorch documentation: \url{https://docs.pytorch.org/docs/stable/generated/torch.nn.Softplus.html}. 

Other constrained parameters could be handled via similarly appropriate mappings.
\subsection{Extension to Minibatches}

For learning on large datasets, as in Case Study B, we need to perform gradient updates on minibatches of a few examples at a time, rather than all $N$ examples in the training set. We approximate the sum over all data in Lines 5 and 6 of Alg.~\ref{alg:deelbo} with a rescaled sum over all items in one minibatch $\mathcal{B}$:
\begin{align}
    \sum_{i=1}^{N} \log p_{\eta}(y_i | \theta_s) \approx \frac{N}{|\mathcal{B}|} \sum_{i \in \mathcal{B}} \log p_{\eta}(y_i | \theta_s).
\end{align}

\addtocounter{section}{-3}
\section{Appendix for Case Study A: Fourier Regression}
\label{app:caseA}
\subsection{RFFs with Arbitrary Lengthscale and Outputscale for Model A}
\label{sec:caseA_lengthscale_and_outputscale}
Random Fourier features (RFFs) are typically used to approximate an RBF kernel with fixed lengthscale $\ell_k = 1$ and output scale $\sigma_k = 1$. Below we show that our RFF construction in Sec.~\ref{sec:caseA_model_definition} allows a Monte Carlo approximation of the RBF kernel for any $\ell_k > 0, \sigma_k > 0$.
Our proof extends a proof for the base case ($\ell_k = 1,\sigma_k = 1$) found in a blogpost \emph{Random Fourier Features} by Gregory Gundersen: \url{https://gregorygundersen.com/blog/2019/12/23/random-fourier-features/}, and likely informed work on RFFs for latent variable models in \citet{gundersen2021latent}.

First, recall basic notation and dimensionality assumptions. Each original feature vector is $x \in \mathbb{R}^H$. At algorithm startup, we draw weights $A \in \mathbb{R}^{H \times R}$ from $A_{h,r} \sim \mathcal{N}(0, 1)$ for all $h \in [H]$ and all $r \in [R]$. Similarly, we draw bias weights $b \in \mathbb{R}^R$ from $b_{r} \sim \operation{Unif}(0, 2\pi)$ for all $r \in [R]$. We typically assume the user-controlled size of the RFF feature space $R$ is rather large. As $R \rightarrow \infty$, the approximate equalities marked by $\approx$ below become increasingly accurate (in expectation).

For any pair of original feature vectors $x, x'$, after applying our RFF transformation $\phi(\cdot)$ in Eq.~\eqref{eq:phi} the inner product of the pair in the transformed space is:
\begin{align}
    \phi(x)^\top\phi(x') 
    &= \frac{\sigma_k^2}{R} \sum_{r=1}^R 2\cos \left( \frac{1}{\ell_k} A_{:,r}^\top x + b_r \right) \cos \left( \frac{1}{\ell_k} A_{:^\top,r}^\top x' + b_r \right) 
    && \text{By definition of $\phi(\cdot)$}
    \\
    &= \frac{\sigma_k^2}{R} \sum_{r=1}^R \cos \left( \frac{1}{\ell_k} A_{:,r}^\top (x+x') + 2b_r \right) + \cos \left( \frac{1}{\ell_k} A_{:^\top,r}^\top (x-x') \right) && \text{Sum of angles formula} \\
    &\approx \frac{\sigma_k^2}{R} \sum_{r=1}^R \cos \left( \frac{1}{\ell_k} A_{:,r}^\top (x-x') \right) && \text{\makecell[lt]{$\mathbb{E}[\cos(t+b)] = 0$ for any\\uniform r.v. $b$ with a $2\pi$\\interval length and any\\scalar $t$}} \\
    &\approx \frac{\sigma_k^2}{R} \sum_{r=1}^R \cos \left( \frac{1}{\ell_k} A_{:,r}^\top (x-x') \right) + i\sin \left( \frac{1}{\ell_k} A_{:^\top,r}^\top (x-x') \right) && \text{\makecell[lt]{$\mathbb{E}[\sin(a)] = 0$ for any zero-\\mean Gaussian r.v. $a$}} \\
    &\approx \frac{\sigma_k^2}{R} \sum_{r=1}^R \exp \left( i \frac{1}{\ell_k} A_{:,r}^\top (x-x') \right) && \text{Euler's formula}
\end{align}
The above sum over $R$ samples can be viewed as an unbiased estimate of an expectation of a complex exponential:
\begin{align}
\label{eq:sumoverr_as_expectation}
    \frac{\sigma_k^2}{R} \sum_{r=1}^R \exp \left( i \frac{1}{\ell_k} A_{:,r}^\top (x-x') \right) \approx \sigma_k^2\mathbb{E}_{p(\omega)} \left[ \exp(i\omega^\top (x-x'))
    \right], &&  \text{where $p(\omega) = 
    \mathcal{N}(\omega | 0_H, \tfrac{1}{\ell_k^{2}} I_H)$}.
\end{align}
Here, the random variable is a vector $\omega \in \mathbb{R}^H$, where each entry is distributed as $\omega_h \sim \mathcal{N}(0, \frac{1}{\ell_k^2})$ for all $h \in [H]$.
Recall that one way to sample values of $\omega$ is in two steps: first draw vector $A_{:,r} \in \mathbb{R}^H$ from a standard zero-mean, identity-covariance Normal in $H$ dimensions (which matches our RFF procedure for $A$), then set $\omega \gets \frac{1}{\ell_k} A_{:,r}$.
This two-step sampling procedure justifies the approximation in Eq.~\eqref{eq:sumoverr_as_expectation}.

Now, we show that the expectation of the complex exponential on the right-side of Eq.~\eqref{eq:sumoverr_as_expectation} is equivalent to the stationary RBF kernel $k_{\textnormal{RBF},\ell_k, \sigma_k}(\delta)$ where $\delta = x - x'$ is an $H$-dimensional vector. The lengthscale $\ell_k$ and outputscale $\sigma_k$ values of the kernel match those used to construct the RFF features $\phi(\cdot)$ above.

$\sigma_k^2\mathbb{E}_{p(\omega)}[\exp(i\omega^\top\delta)]$
\begin{align}
    & = \sigma_k^2 \int p\left(\omega\right) \exp\left(i\omega^\top\delta\right) d\omega 
    && \text{Expectation as integral}
    \\
    &= \sigma_k^2 \int \left(\frac{\ell_k^2}{2\pi}\right)^{H/2} \exp\left(- \frac{\ell_k^2\omega^\top\omega}{2} \right) \exp\left(i\omega^\top\delta\right) d\omega 
    && \text{By definition of $p(\omega)$}
    \\
    &= \sigma_k^2 \left(\frac{\ell_k^2}{2\pi}\right)^{H/2} \int \exp\left(- \frac{\ell_k^2\omega^\top\omega}{2} + i\omega^\top\delta\right) d\omega
    && \text{Simplify}
    \\
    &= \sigma_k^2 \left(\frac{\ell_k^2}{2\pi}\right)^{H/2} \int \exp\left(- \frac{\ell_k^2\omega^\top\omega}{2} + i\omega^\top\delta + \frac{\delta^\top\delta}{2\ell_k^2} - \frac{\delta^\top\delta}{2\ell_k^2}\right) d\omega 
    && \text{Add and subtract same term}
    \\ 
    &= \sigma_k^2 \left(\frac{\ell_k^2}{2\pi}\right)^{H/2} \exp\left( - \frac{\delta^\top\delta}{2\ell_k^2} \right) \int \exp\left(- \frac{\ell_k^2}{2} \left(\omega - \frac{i}{\ell_k^2}\delta\right)^\top \left(\omega - \frac{i}{\ell_k^2}\delta\right) \right) d\omega 
    && \text{Complete the square}
    \\
    &= 
    \sigma_k^2 \left(\frac{\ell_k^2}{2\pi}\right)^{H/2} \exp\left( - \frac{\delta^\top\delta}{2\ell_k^2} \right) \left(\frac{2\pi}{\ell_k^2}\right)^{H/2} 
    && \text{Closed-form integral}
    \\
    &= \sigma_k^2 \exp\left( - \frac{\delta^\top\delta}{2\ell_k^2} \right) && \text{Reciprocal terms cancel out}
    \\
    &= k_{\textnormal{RBF},\ell_k,\sigma_k}(\delta)
    && \text{By definition of RBF kernel}
\end{align}
This completes the proof. Our constructed RFF features with arbitrary lengthscale and outputscale in Eq.~\eqref{eq:phi} provide an unbiased approximation of the corresponding RBF kernel that is increasingly accurate as $R \rightarrow \infty$.
\subsection{Regression Model A Definition}
\label{sec:caseA_regression_model}

A Bayesian interpretation of the RFF problem assumes a joint probabilistic model $p(y_{1:N}, v) = p(v) p(y_{1:N} | v)$, with factors for the regression case
\begin{align}
    p(v) = \mathcal{N}(v | 0_R, I_R), \qquad p(y_{1:N} | v) = \mathcal{N}(y_{1:N} | \Phi v, \sigma_y^2 I_N)
\end{align}
where $y_{1:N} \in \mathbb{R}^N$ and $\Phi = \Phi(x_{1:N}) \in \mathbb{R}^{N \times R}$.
Note that including a prior variance hyperparameter different from 1 for $p(v)$ does not add an additional degree of freedom when $\Phi$ includes an outputscale hyperparameter. Both factors would be redundant in the inner product $v^\top \phi(x_i)$.

\textbf{Marginal likelihood.}
Given the marginal Gaussian distribution for $v$ and the conditional Gaussian distribution for $y_{1:N}$ given $v$, the marginal distribution of $y_{1:N}$ under our RFF regression model is:
\begin{align}
    p(y_{1:N}) &= \mathcal{N}(y_{1:N} | 0_N, \sigma_y^2 I_N + \Phi \Phi^\top) && \text{by Eq.~(2.115) from \cite{bishop2006pattern}}.
\end{align}
This is \emph{exactly the same} marginal likelihood as the classic Gaussian process with RBF kernel regression model, where we have latent function values $f \sim \mathcal{N}(0_N, K)$ and then observed targets $y | f \sim \mathcal{N}(f, \sigma_y^2 I_N)$, so long as the approximation $K \approx \Phi \Phi^\top$ is accurate.

\textbf{Ideal posterior.} The conditional distribution of $v$ given $y$ is available in closed form:
\begin{align}
    p(v | y_{1:N}) &= \mathcal{N}(v | \tfrac{1}{\sigma_y^2} \Sigma_{\text{post}} \Phi^\top y_{1:N}, \Sigma_{\text{post}}) && \text{by Eq.~(2.116) from \cite{bishop2006pattern}} \\
    \intertext{where}
    \Sigma_{\text{post}} &= (I_R + \tfrac{1}{\sigma_y^2} \Phi^\top \Phi)^{-1} && \text{by Eq.~(2.117) from \cite{bishop2006pattern}}.
\end{align}

\begin{figure}[htbp!]
    \centering
    \includegraphics[width=195.129pt]{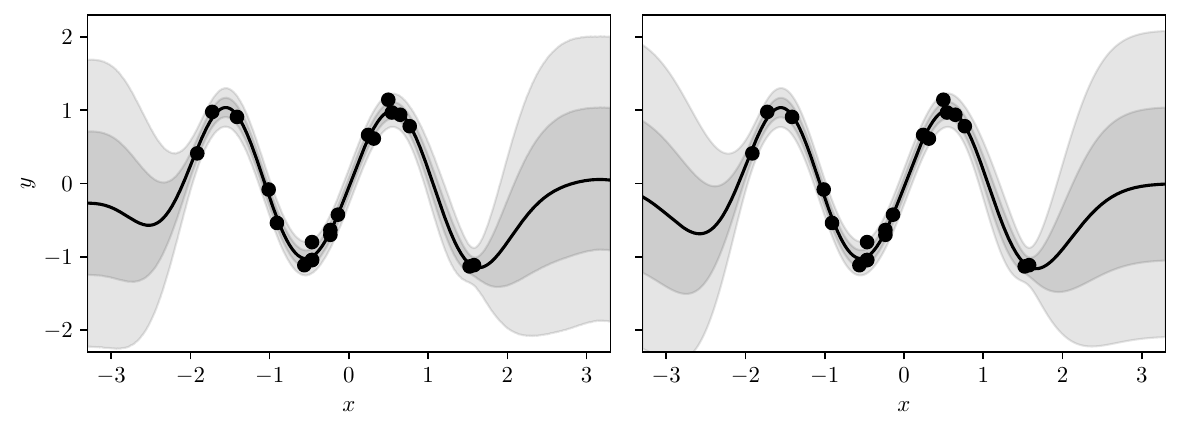} \\
    \begin{subfigure}{0.1951\linewidth}
        \captionsetup{font=scriptsize,labelfont=scriptsize}
        \subcaption{\makecell{RFF (learned $\ell_k, \sigma_k$)\\LML}}
    \end{subfigure}
    \begin{subfigure}{0.1951\linewidth}
        \captionsetup{font=scriptsize,labelfont=scriptsize}
        \subcaption{\makecell{GP (learned $\ell_k, \sigma_k$)\\LML}}
    \end{subfigure}
    \caption{Left: A Monte Carlo approximation of the RFF predictive posterior $p(y_* | y_{1:N}) = \int_v p(y_* | v) p(v | y_{1:N}) dv$ by sampling from the true posterior $p(v | y_{1:N})$. Right: The closed-form GP predictive posterior $p(y_* | y_{1:N})$.}
    \label{fig:rff_gp_comparison}
\end{figure}

The posterior predictive $p(y_* | y_{1:N})$ of the RFF regression model again should match the GP with corresponding RBF kernel, provided the user-controlled rank $R$ is large enough. See Fig.~\ref{fig:rff_gp_comparison} for a visual demonstration.

\subsection{Closed-Form Optimal \texorpdfstring{$\psi$}{Variational Parameters} for Full-Rank \texorpdfstring{$q$}{Approximate Posterior} Under ELBO for Regression Model A}
\label{sec:caseA_closed-form_elbo}

Suppose we assume a full-rank covariance for $q$, rather than isotropic: $q(v) = \mathcal{N}(v | \mu_q, \Sigma_q)$. In this case, the closed-form ELBO for regression Model A is
\begin{align}
    J_{\text{ELBO}} = - \frac{1}{2} \left[ N \log (2\pi\sigma_y^2) + \frac{1}{\sigma_y^2} \| y_{1:N} - \Phi \mu_q \|_2^2 + \frac{1}{\sigma_y^2} \operation{tr}(\Phi \Sigma_q \Phi^\top) + \operation{tr}(\Sigma_q) + \| \mu_q \|_2^2 - R - \log \det(\Sigma_q) \right].
\end{align}
We can solve for $\Sigma_q$ by taking the gradient of $J_{\text{ELBO}}$ with respect to $\Sigma_q$. The gradient is
\begin{align}
    \nabla_{\Sigma_q} J_{\text{ELBO}} = - \frac{1}{2} \left[ \frac{1}{\sigma_y^2} \Phi^\top \Phi + I_R - \Sigma_q^{-1} \right].
\end{align}
Setting $\nabla_{\Sigma_q} J_{\text{ELBO}} = 0$ and solving for $\Sigma_q$, we get 
\begin{align}
    \Sigma_q^* &= (I_R + \tfrac{1}{\sigma_y^2} \Phi^\top \Phi)^{-1}
\end{align}
which is exactly the ideal posterior's covariance matrix.

\subsection{Closed-Form Optimal \texorpdfstring{$\psi$}{Variational Parameters} for Isotropic \texorpdfstring{$q$}{Approximate Posterior} Under ELBO for Regression Model A}
\label{sec:caseA_closed-form_elbo_isotropic_q}
If we assume $\Sigma_q = \bar{\sigma}_q^2 I_R$, we can simplify the ELBO to
\begin{align}
    J_{\text{ELBO}} = - \frac{1}{2} \left[ N \log (2\pi\sigma_y^2) + \frac{1}{\sigma_y^2} \| y_{1:N} - \Phi \mu_q \|_2^2 + \frac{\bar{\sigma}_q^2}{\sigma_y^2} \operation{tr}(\Phi \Phi^\top) + \bar{\sigma}_q^2 R + \| \mu_q \|_2^2 - R - \log \bar{\sigma}_q^{2R} \right].
\end{align}
We can solve for $\bar{\sigma}_q^2$ by taking the gradient of $J_{\text{ELBO}}$ with respect to $\bar{\sigma}_q^2$. The gradient is
\begin{align}
    \nabla_{\bar{\sigma}_q^2} J_{\text{ELBO}} = -\frac{1}{2} \left[ \frac{1}{\sigma_y^2} \operation{tr}(\Phi \Phi^\top) + R - \frac{1}{\bar{\sigma}_q^2} R \right].
\end{align}
Setting $\nabla_{\bar{\sigma}_q^2} J_{\text{ELBO}} = 0$ and solving for $\bar{\sigma}_q^2$, we get 
\begin{align}
    \bar{\sigma}_q^{2*} = \frac{R}{\frac{1}{\sigma_y^2}\operation{tr}(\Phi\Phi^\top) + R}.
\end{align}

\subsection{Closed-Form Optimal \texorpdfstring{$\psi$}{Variational Parameters} for Isotropic \texorpdfstring{$q$}{Approximate Posterior} Under DE-ELBO for Regression Model A}
\label{sec:caseA_closed-form_deelbo_isotropic_q}
Here we again assume an isotropic covariance for $q$, but now focus on the DE-ELBO.
We can solve for $\bar{\sigma}_q^2$ by taking the gradient of $J_{\text{DE-ELBO}}$ ($\kappa = \frac{R}{N}$) with respect to $\bar{\sigma}_q^2$. The gradient is
\begin{align}
    \nabla_{\bar{\sigma}_q^2} J_{\text{DE-ELBO}} = -\frac{1}{2} \left[ \frac{R}{N} \frac{1}{\sigma_y^2} \operation{tr}(\Phi \Phi^\top) + R - \frac{1}{\bar{\sigma}_q^2} R \right].
\end{align}
Setting $\nabla_{\bar{\sigma}_q^2} J_{\text{DE-ELBO}} = 0$ and solving for $\bar{\sigma}_q^2$, we get 
\begin{align}
    \bar{\sigma}_q^{2*} = \frac{R}{\frac{R}{N} \frac{1}{\sigma_y^2}\operation{tr}(\Phi\Phi^\top) + R}.
\end{align}

\subsection{Classification Model A Definition}
\label{sec:caseA_classification_model}

Building on the regression model above, we now  consider an RFF classifier for when observed data are discrete labels $y_i \in \{1, 2, \ldots, C\}$.
A Bayesian interpretation of this problem assumes a joint probabilistic model $p(y_{1:N}, V) = p(V) \prod_{i} p(y_i | V)$, with factors for the classification case
\begin{align}
    \label{eq:rff_joint_pdf_classification_model}
    p(V) = \mathcal{N}( \operation{vec}(V) | 0_{RC}, I_{RC}), \qquad p(y_i | V) = \operation{Cat}( y_i | \textsc{sm}( V\phi(x_i) ) ).
\end{align}
Unlike the regression case, there is no known analytical formula for the posterior, due to the non-linear softmax function preventing conjugacy.

\subsection{Results for Model A: Regression and Classification}
\begin{figure*}[!htbp]
    \begin{center}
        \includegraphics[width=\linewidth]{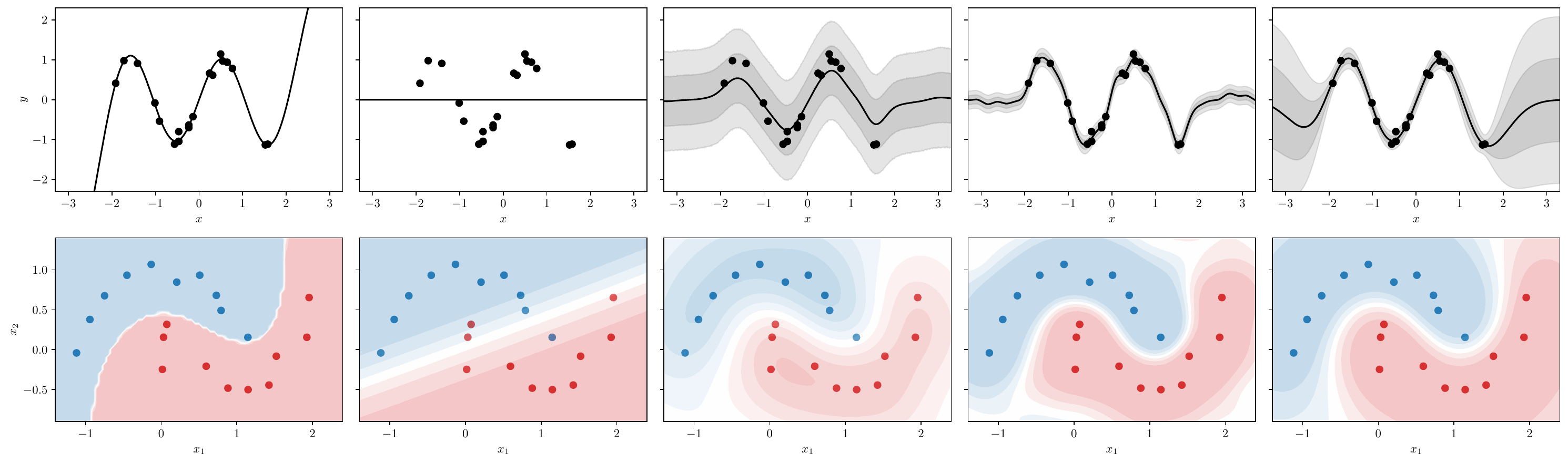}
        \begin{subfigure}{0.1951\linewidth}
            \captionsetup{font=scriptsize,labelfont=scriptsize}
            \subcaption{\makecell{RFF ($\ell_k = 1$)\\MAP + GS}}
        \end{subfigure}
        \begin{subfigure}{0.1951\linewidth}
            \captionsetup{font=scriptsize,labelfont=scriptsize}
            \subcaption{\makecell{RFF ($\ell_k = 20$)\\MAP + GS}}
        \end{subfigure}
        \begin{subfigure}{0.1951\linewidth}
            \captionsetup{font=scriptsize,labelfont=scriptsize}
            \subcaption{\makecell{RFF (learned $\ell_k,\sigma_k$)\\ELBO}}
        \end{subfigure}
        \begin{subfigure}{0.1951\linewidth}
            \captionsetup{font=scriptsize,labelfont=scriptsize}
            \subcaption{\makecell{RFF (learned $\ell_k,\sigma_k$)\\DE-ELBO (ours)}}
        \end{subfigure}
        \begin{subfigure}{0.1951\linewidth}
            \captionsetup{font=scriptsize,labelfont=scriptsize}
            \subcaption{\makecell{GP (learned $\ell_k,\sigma_k$)\\LML}}
        \end{subfigure}
    \end{center}
    \caption{Demo of hyperparameter sensitivity and selection for RFF models.
    The first four columns use the RFF regression model with isotropic Gaussian $q$ in Sec.~\ref{sec:caseA_model_definition}, varying estimation and selection techniques.
    The last column shows the reference fit of a GP's exact posterior, a gold-standard for this toy data but less scalable.
    For regression, we plot the mean and two standard deviations for the predictive posterior $p(y_* | y_{1:N})$.
    Our DE-ELBO objective best approximates the GP, though underestimates variance far from data.
    }
    \label{fig:caseA_regression_classification_demo}
\end{figure*}

\newpage
\subsection{Varying \texorpdfstring{$\kappa$}{Data-Emphasis Factor} for Model A}
\label{sec:caseA_varying_kappa}

\begin{figure}[htbp!]
    \centering
    \includegraphics[width=\linewidth]{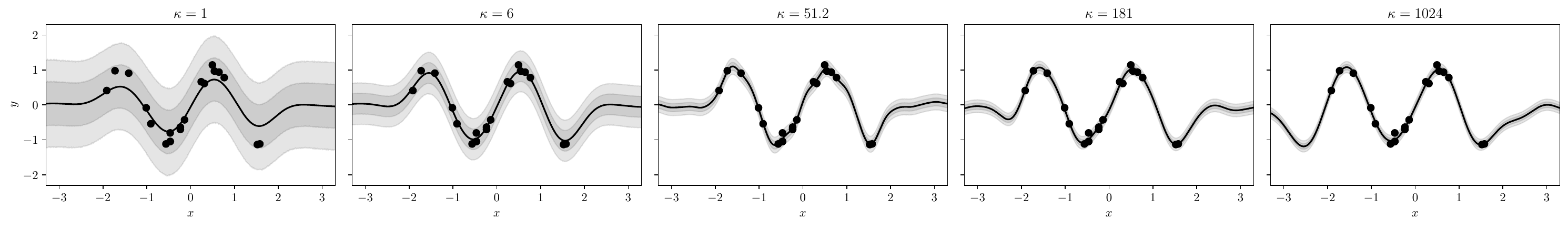}
    \caption{Predictive posteriors on a univariate regression dataset using our \emph{data-emphasized ELBO} (DE-ELBO) with various $\kappa$ values ($\frac{D}{N} = 51.2$).
    We use Adam with the closed-form expected log-likelihood.
    The predictive posterior is slightly better compared to Adam with a Monte Carlo approximation of the expected log-likelihood (see Fig.~\ref{fig:caseA_regression_demo}~(c)).}
    \label{fig:rff_varying_kappa}
\end{figure}

\section{Appendix for Case Study B: Transfer Learning}
\label{app:caseB}
\subsection{Implementation Details for Model B}
\label{sec:caseB_implementation_details}

For MAP + GS, we select the initial learning rate from \{0.1, 0.01, 0.001, 0.0001\} and $\alpha,\beta$ (or equivalently $\lambda,\tau$). 
For L2-zero, we select $\frac{\alpha}{N} = \frac{\beta}{N}$ from \{0.01, 0.001, 0.0001, 1e-5, 1e-6, 0.0\}.
For L2-SP, we select $\frac{\alpha}{N}$ from \{0.01, 0.001, 0.0001, 1e-5, 1e-6, 0.0\} and $\frac{\beta}{N}$ from \{0.01, 0.001, 0.0001, 1e-5, 1e-6, 0.0\}.
For PTYL, we select $\lambda$ from \{1, 10, 100, 1000, 10000, 1e5, 1e6, 1e7, 1e8, 1e9\} and $\frac{\beta}{N}$ from \{0.01, 0.001, 0.0001, 1e-5, 1e-6, 0.0\}.

For MAP + BO~\citep{hvarfner2024vanilla}, we select the initial learning rate from [0.1, 0.0001] and $\alpha,\beta$ (or equivalently $\lambda,\tau$). For L2-SP, we select $\frac{\alpha}{N}$ from [0.01, 1e-6] and $\frac{\beta}{N}$ from [0.01, 1e-6].

For diagEF LA-LML~\citep{immer2021scalable}, we select the initial learning rate from \{0.1, 0.01, 0.001, 0.0001\} and learn $\alpha,\beta$ (or equivalently $\lambda,\tau$).

For diagEF LA-CLML~\citep{lotfi2022bayesian}, we select the initial learning rate from \{0.1, 0.01, 0.001, 0.0001\} and $\alpha,\beta$ (or equivalently $\lambda,\tau$).
For L2-SP, we select $\frac{\alpha}{N}$ from \{0.01, 0.001, 0.0001, 1e-5, 1e-6, 0.0\} and $\frac{\beta}{N}$ from \{0.01, 0.001, 0.0001, 1e-5, 1e-6, 0.0\}.

For iso ELBO and iso DE-ELBO (ours), we select the initial learning rate from \{0.1, 0.01, 0.001, 0.0001\} and learn $\lambda, \tau$.

\subsection{Closed-Form Optimal \texorpdfstring{$\eta$}{Hyperparameters} for Model B}
\label{sec:caseB_learning_lambda_tau}

In our particular model in Eq.~\eqref{eq:caseB_joint_pdf}, the KL divergence between two Gaussians \citep{murphy2022example} simplifies for the backbone KL term as:
\begin{align}
-D_{\text{KL}}(q(w) \| p(w)) = -\frac{1}{2} \left[ \frac{\bar{\sigma}_q^2}{\lambda} \operation{tr} (\Sigma_p^{-1}) + \frac{1}{\lambda} (\mu_p-\bar{w})^\top\Sigma_p^{-1}(\mu_p-\bar{w}) - F + \log \left( \frac{\lambda^{F}\det(\Sigma_p)}{\bar{\sigma}_q^{2F}} \right) \right].
\end{align}

\textbf{Closed-form updates.} To find an optimal $\lambda$ value with respect to the $J_{\text{ELBO}}$, notice that of the 3 additive terms in Eq.~\eqref{eq:de_elbo_objective}, only the KL term between $q(w)$ and $p(w)$ involves $\lambda$. We solve for $\lambda$ by taking the gradient of the KL term with respect to $\lambda$, setting to zero, and solving, with assurances of a local maximum of $J_{\text{ELBO}}$ via a second derivative test. The gradient is
\begin{align}
    \nabla_\lambda -D_{\text{KL}}(q(w) \| p(w)) = -\frac{1}{2} \left[ - \frac{\bar{\sigma}_q^2}{\lambda^2} \operation{tr}(\Sigma_p^{-1}) - \frac{1}{\lambda^2} (\mu_p-\bar{w})^\top \Sigma_p^{-1} (\mu_p-\bar{w}) + \frac{F}{\lambda} \right].
\end{align}
Setting $\nabla_\lambda -D_{\text{KL}}(q(w) \| p(w)) = 0$ and solving for $\lambda$, we get 
\begin{align}
    \lambda^* = \frac{1}{F} \Big[ \bar{\sigma}_q^2 \operation{tr}(\Sigma_p^{-1}) + (\mu_p-\bar{w})^\top \Sigma_p^{-1} (\mu_p-\bar{w}) \Big].    
\end{align}

In our particular model in Eq.~(\ref{eq:caseB_joint_pdf}), the KL divergence between two Gaussians \citep{murphy2022example} simplifies for the classifier head KL term as:
\begin{align}
    -D_{\text{KL}}(q(V) \| p(V)) = -\frac{1}{2} \left[ \frac{\bar{\sigma}_q^2}{\tau} HC + \frac{1}{\tau} || \operation{vec}(\bar{V}) ||_2^2 - HC + \log \left( \frac{\tau^{HC}}{\bar{\sigma}_q^{2HC}} \right)\right].
\end{align}
To find an optimal $\tau$ value with respect to the $J_{\text{ELBO}}$, notice that of the 3 additive terms in Eq.~\eqref{eq:de_elbo_objective}, only the KL term between $q(V)$ and $p(V)$ involves $\tau$. We solve for $\tau$ by taking the gradient of the KL term with respect to $\tau$, setting to zero, and solving, with assurances of a local maximum of $J_{\text{ELBO}}$ via a second derivative test. The gradient is
\begin{align}
    \nabla_\tau -D_{\text{KL}}(q(V) \| p(V)) = -\frac{1}{2} \left[ - \frac{\bar{\sigma}_q^2}{\tau^2} HC - \frac{1}{\tau^2} || \operation{vec}(\bar{V}) ||_2^2 + \frac{1}{\tau} HC \right].
\end{align}
Setting $\nabla_\tau -D_{\text{KL}}(q(V) \| p(V)) = 0$ and solving for $\tau$, we get 
\begin{align}
    \tau^* = \frac{\bar{\sigma}_q^2 HC + || \operation{vec}(\bar{V}) ||_2^2}{HC}.
\end{align}

\textbf{Second derivative tests.}
To verify the optima found above, we perform second derivative tests.
The second derivative of the negative KL term with respect to $\lambda$ is:
\begin{align}
    \nabla^2_\lambda -D_{\text{KL}}(q(w) \| p(w)) &= -\frac{1}{2} \left[ \frac{2\bar{\sigma}_q^2}{\lambda^3} \operation{tr}(\Sigma_p^{-1}) + \frac{2}{\lambda^3} (\mu_p - \bar{w})^\top \Sigma_p^{-1} (\mu_p - \bar{w}) - \frac{F}{\lambda^2} \right] \\
    &= -\frac{1}{2} \left[ \frac{2F}{\lambda^3} \frac{1}{F} \left( \bar{\sigma}_q^2 \operation{tr}(\Sigma_p^{-1}) + (\mu_p - \bar{w})^\top \Sigma_p^{-1} (\mu_p - \bar{w}) \right) - \frac{F}{\lambda^2} \right] \\
    &= -\frac{1}{2} \left[ \frac{2F}{\lambda^3} \lambda^* - \frac{F}{\lambda^2} \right].
\end{align}
Plugging in $\lambda^*$ and simplifying, we get
\begin{align}
    \nabla^2_\lambda -D_{\text{KL}}(q(w) \| p(w | \lambda^*)) &= -\frac{F}{2}  \frac{1}{\lambda^{*2}}
\end{align}
This expression is always negative, indicating that $\lambda^*$ is a local maximum of $J_{\text{ELBO}}$.

The second derivative of the negative KL term with respect to $\tau$ is:
\begin{align}
    \nabla_\tau^2 -D_{\text{KL}}(q(V) \| p(V)) &= -\frac{1}{2} \left[ \frac{2\bar{\sigma}_q^2}{\tau^3} HC + \frac{2}{\tau^3} || \operation{vec}(\bar{V}) ||_2^2 - \frac{1}{\tau^2} HC \right] \\
    &= -\frac{1}{2} \left[  \frac{2HC}{\tau^3}  \left( \bar{\sigma}_q^2 + \frac{1}{HC} || \operation{vec}(\bar{V}) ||_2^2 \right) - \frac{HC}{\tau^2} \right] \\ 
    &= -\frac{1}{2} \left[  \frac{2HC}{\tau^3} \tau^*  - \frac{HC}{\tau^2} \right].
\end{align}
Plugging in $\tau^*$ and simplifying, we get
\begin{align}
    \nabla^2_\tau -D_{\text{KL}}(q(V) \| p(V | \tau^*)) &= -\frac{HC}{2}  \frac{1}{\tau^{*2}}.
\end{align}
This expression is always negative, indicating that $\tau^*$ is a local maximum of $J_{\text{ELBO}}$.

\subsection{Low-rank \texorpdfstring{$\Sigma_p$}{Prior Covariance Matrix} for Model B}
\label{sec:caseB_low-rank_Sigma_p}

The PTYL method \citep{shwartz2022pre} uses Stochastic Weight Averaging-Gaussian (SWAG;~\citealp{maddox2019simple}) to approximate the posterior distribution $p(w|\mathcal{D}_S)$ of the source data $\mathcal{D}_S$ with a Gaussian distribution $\mathcal{N}(\mu, \Sigma)$ where $\mu$ is the learned mean and $\Sigma = \frac{1}{2}(\Sigma_{\text{diag}} + \Sigma_{\text{LR}})$ is a representation of a covariance matrix with both diagonal and \emph{low-rank} components.
The LR covariance has the form $\Sigma_{\textrm{LR}} = \frac{1}{K-1} Q Q^\top$, where $Q \in \mathbb{R}^{F \times K}$.

We use the Woodbury matrix identity \citep{woodbury1950inverting}, trace properties, and the matrix determinant lemma to compute the trace of the inverse, squared Mahalanobis distance, and log determinant of the low-rank covariance matrix for the KL term.

The trace and log determinant of the low-rank covariance matrix can be calculated once and used during training.
Like in the PTYL method, the squared Mahalanobis distance needs to be re-evaluated every iteration of gradient descent.

\textbf{Trace of the inverse.}
We compute the trace of the inverse of the low-rank covariance matrix using the Woodbury matrix identity and trace properties.
\begin{align*}
\operation{tr} (\Sigma_p^{-1}) &= \operation{tr}( (A + UCV )^{-1} ) \\
&= \operation{tr} (A^{-1} - A^{-1}U(C^{-1} + VA^{-1}U)^{-1}VA^{-1}) && \text{Woodbury matrix identity} \\
&= \operation{tr} (A^{-1}) - \operation{tr}(A^{-1}U(C^{-1} + VA^{-1}U)^{-1}VA^{-1}) && \operation{tr}(A+B) = \operation{tr}(A) + \operation{tr}(B) \\
&= \operation{tr} (A^{-1}) - \operation{tr}((C^{-1} + VA^{-1}U)^{-1}VA^{-1}A^{-1}U) && \operation{tr}(AB) = \operation{tr}(BA)
\end{align*}
where $A=\frac{1}{2}\Sigma_{\text{diag}}$, $C = I_K$, $U = \frac{1}{\sqrt{2K-2}}Q$, and $V=\frac{1}{\sqrt{2K-2}}Q^\top$.
The last trace property, lets us compute the trace of the inverse of the low-rank covariance matrix without having to store a $F \times F$ covariance matrix.

\textbf{Squared Mahalanobis distance.}
We compute the squared Mahalanobis distance $(\mu_p-\bar{w})^\top\Sigma_p^{-1}(\mu_p-\bar{w})$ by distributing the mean difference vector into the Woodbury matrix identity.
\begin{align*}
\Sigma_p^{-1} &= (A + UCV )^{-1} \\
&= (A^{-1} - A^{-1}U(C^{-1} + VA^{-1}U)^{-1}VA^{-1}) && \text{Woodbury matrix identity}
\end{align*}

\textbf{Log determinant.}
We compute the log determinant of the low-rank covariance matrix using the matrix determinant lemma.
\begin{align*}
\log \det(\Sigma_p) &= \log \det(A + UV) \\
&= \log (\det(I_K + VA^{-1}U) \det(A))  && \text{Matrix determinant lemma}
\end{align*}

\subsection{Varying \texorpdfstring{$\kappa$}{Data-Emphasis Factor} for Model B}
\label{sec:caseB_varying_kappa}

\begin{figure*}[htbp!]
  \centering
  \includegraphics[width=\linewidth]{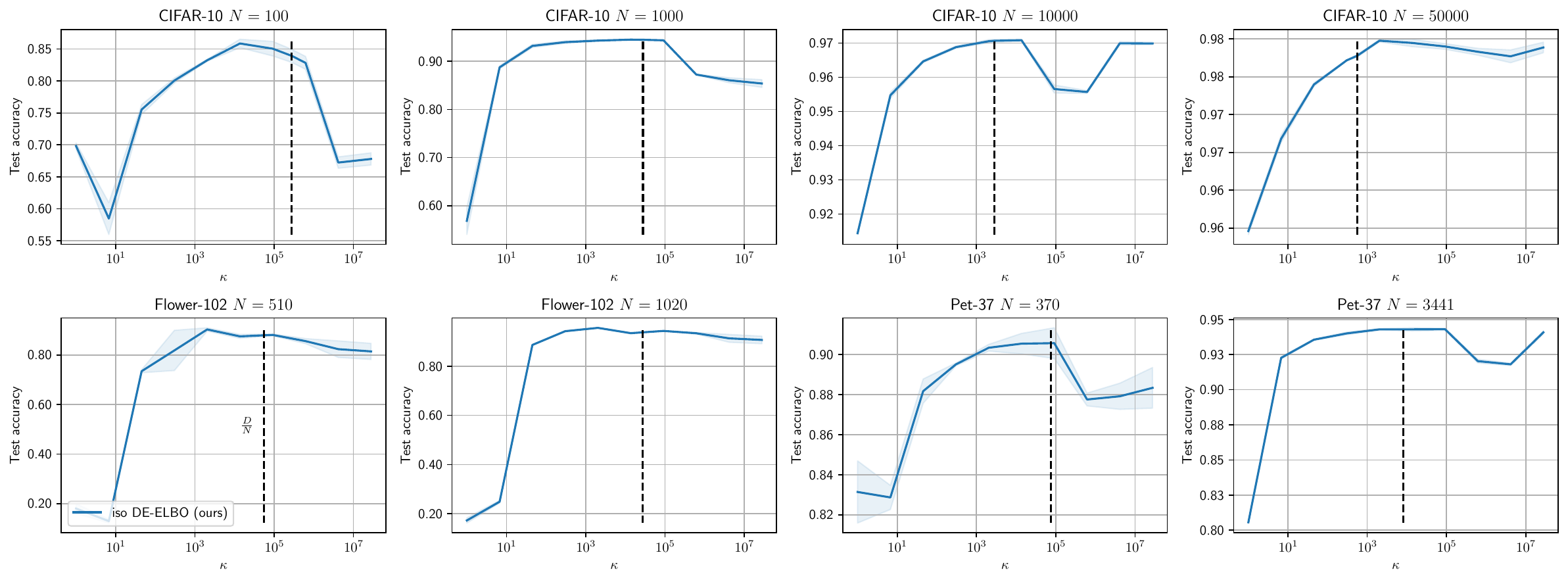}
  \caption{Test set accuracy over $\kappa$ values for L2-SP with iso DE-ELBO (ours). We report the mean (std) over 3 separately-sampled training sets. Each pannel shows a different task: ConvNeXt-Tiny fine-tuned on CIFAR-10, Flower-102, and Pet-37.
  }
  \label{fig:convnext_tiny_varying_kappa_acc}
\end{figure*}
\begin{figure*}[htbp!]
  \centering
  \includegraphics[width=\linewidth]{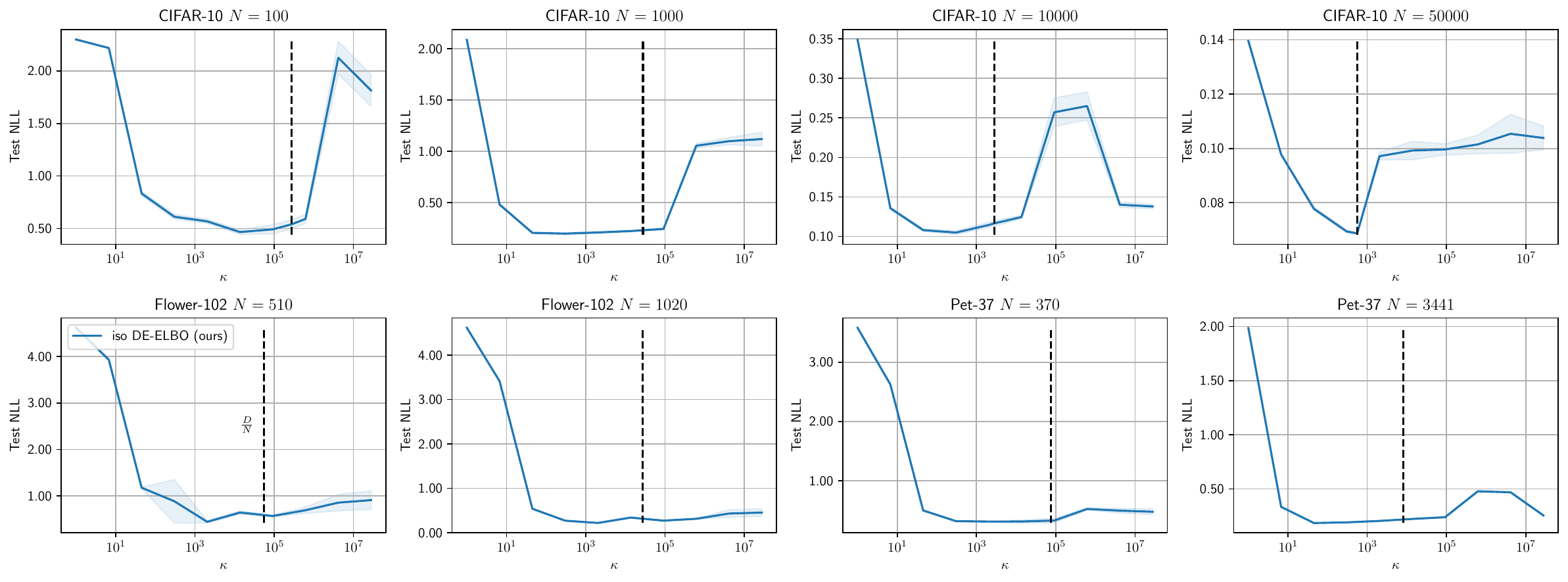}
  \caption{Test set NLL over $\kappa$ values for L2-SP with iso DE-ELBO (ours). We report the mean (std) over 3 separately-sampled training sets. Each pannel shows a different task: ConvNeXt-Tiny fine-tuned on CIFAR-10, Flower-102, and Pet-37.
  }
  \label{fig:convnext_tiny_varying_kappa_nll}
\end{figure*}
\newpage
\subsection{Computational Time Comparisons for Model B}
\label{sec:caseB_computational_time_comparisons}

\begin{figure*}[htbp!]
  \centering
  \includegraphics[width=\linewidth]{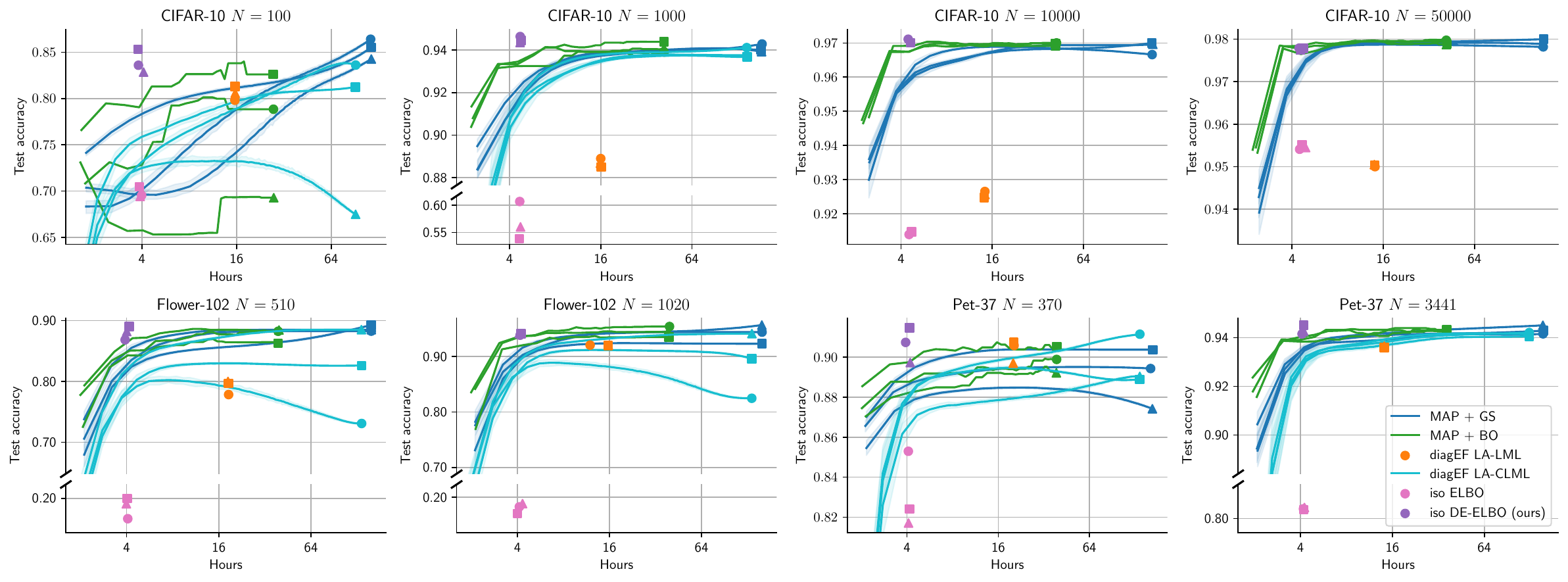}
  \caption{Test set accuracy over time for L2-SP transfer learning methods.
  We run each method on 3 separate training sets of size $N$ (3 different marker styles).
  Each panel shows a different task: ConvNeXt-Tiny fine-tuned on CIFAR-10, Flower-102, and Pet-37.
  We compare MAP + GS, MAP + BO~\citep{hvarfner2024vanilla}, diagEF LA-LML~\citep{immer2021scalable}, diagEF LA-CLML~\citep{lotfi2022bayesian}, iso ELBO, and iso DE-ELBO (ours).
  To make the blue curves, we did the full grid search once (markers). Then, for each grid search size, we subsampled that number of hyperparameter configurations and selected the best using validation NLL. Averaging this over 500 subsamples for each grid size produced the blue lines.
  To make the green curves, we use BO to select candidate hyperparameter configurations and selected the best using validation NLL. Averaging this over 5 BO runs produced the green lines.
  }
  \label{fig:convnext_tiny_computational_time_comparison}
\end{figure*}
\begin{figure*}[htbp!]
  \centering
  \includegraphics[width=\linewidth]{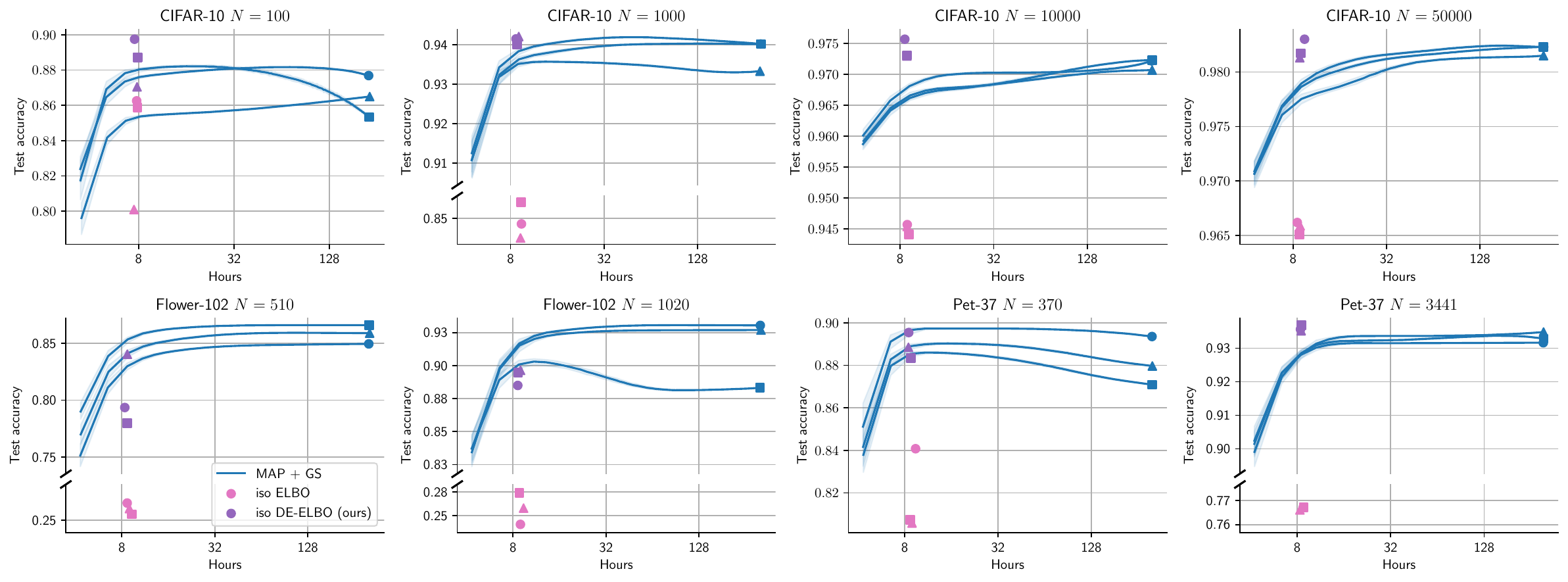}
  \caption{Test set accuracy over time for L2-SP transfer learning methods.
  We run each method on 3 separate training sets of size $N$ (3 different marker styles).
  Each panel shows a different task: ViT-B/16 fine-tuned on CIFAR-10, Flower-102, and Pet-37.
  We compare MAP + GS, iso ELBO, and iso DE-ELBO (ours).
  To make the blue curves, we did the full grid search once (markers). Then, for each grid search size, we subsampled that number of hyperparameter configurations and selected the best using validation NLL. Averaging this over 500 subsamples for each grid size produced the blue lines.
  }
  \label{fig:vit_b_16_computational_time_comparison}
\end{figure*}
\begin{figure*}[htbp!]
  \centering
  \includegraphics[width=\linewidth]{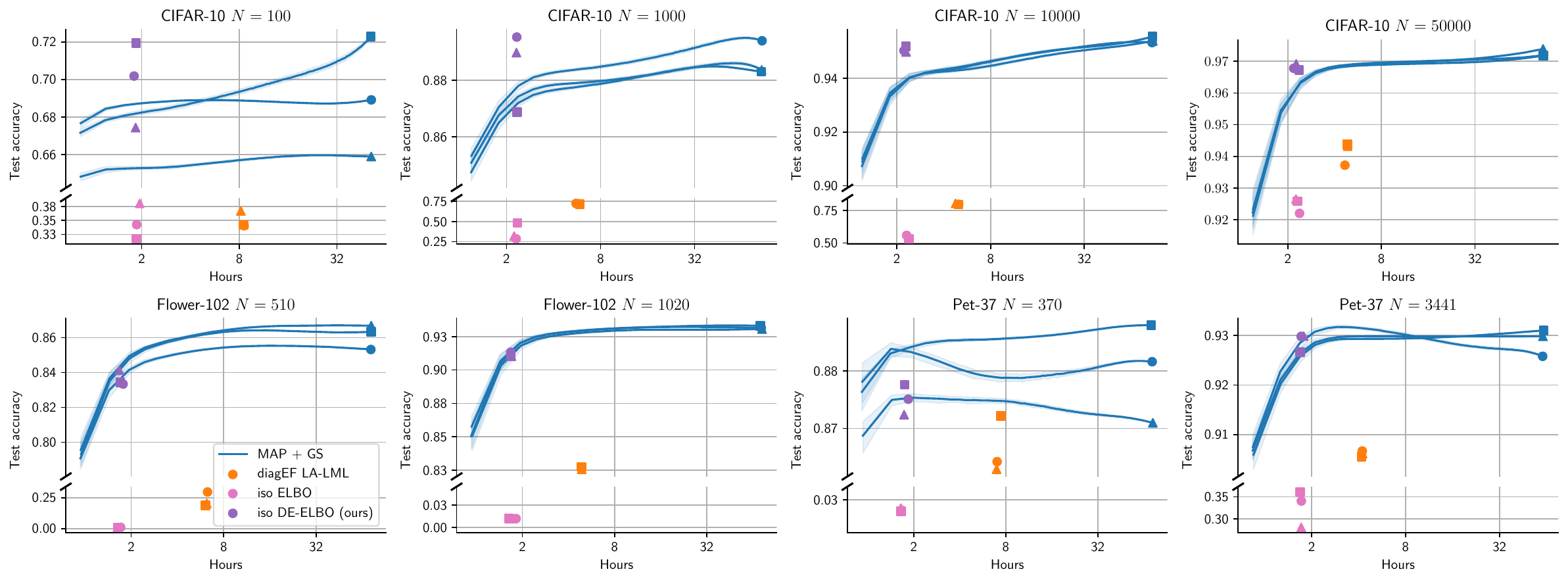}
  \caption{Test set accuracy over time for L2-SP transfer learning methods.
  We run each method on 3 separate training sets of size $N$ (3 different marker styles).
  Each panel shows a different task: ResNet-50 fine-tuned on CIFAR-10, Flower-102, and Pet-37.
  We compare MAP + GS, diagEF LA-LML~\citep{immer2021scalable}, iso ELBO, and iso DE-ELBO (ours).
  To make the blue curves, we did the full grid search once (markers). Then, for each grid search size, we subsampled that number of hyperparameter configurations and selected the best using validation NLL. Averaging this over 500 subsamples for each grid size produced the blue lines.
  }
  \label{fig:resnet_50_computational_time_comparison}
\end{figure*}
\begin{figure*}[t!]
  \centering
  \includegraphics[width=\linewidth]{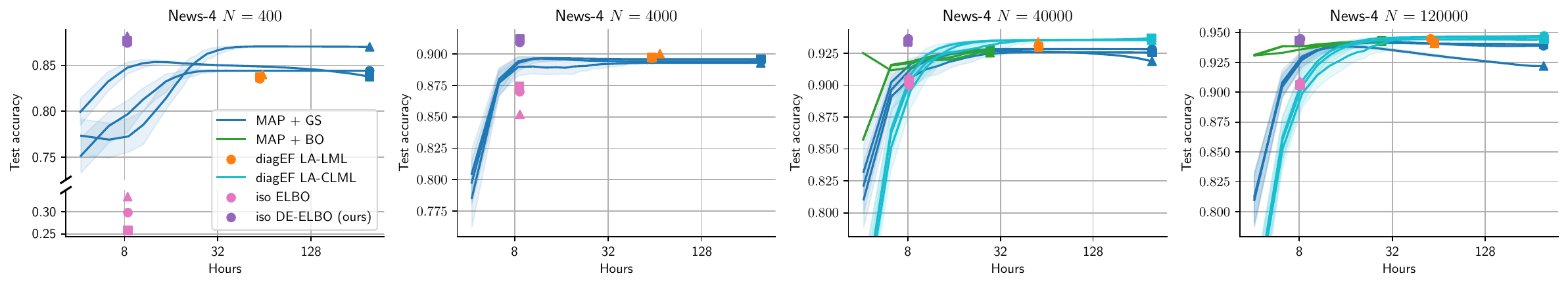}
  \caption{Test set accuracy over time for L2-SP transfer learning methods.
  We run each method on 3 separate training sets of size $N$ (3 different marker styles).
  Task: BERT-base fine-tuned on News-4.
  We compare MAP + GS, MAP + BO~\citep{hvarfner2024vanilla}, diagEF LA-LML~\citep{immer2021scalable}, diagEF LA-CLML~\citep{lotfi2022bayesian}, iso ELBO, and iso DE-ELBO (ours).
  To make the blue curves, we did the full grid search once (markers). Then, for each grid search size, we subsampled that number of hyperparameter configurations and selected the best using validation NLL. Averaging this over 500 subsamples for each grid size produced the blue lines.
  To make the green curves, we use BO to select candidate hyperparameter configurations and selected the best using validation NLL. Averaging this over 10 BO runs produced the green lines.
  }
  \label{fig:bert_base_computational_time_comparison}
\end{figure*}

\newpage
\subsection{Results for Model B}
\label{sec:caseB_results}

\setlength{\tabcolsep}{2pt}
\setcounter{table}{5}
\begin{table*}[htbp!]
  \caption{Accuracy on CIFAR-10, Flower-102, and Pet-37 test sets for different probabilistic models, methods, and backbones. We report the mean (min-max) over 3 separately-sampled training sets. MAP + GS requires 24 different SGD runs for L2-zero, 144 for L2-SP, and 240 for PTYL, while diagEF LA-LML~\citep{immer2021scalable} and iso DE-ELBO (ours) require 4 different SGD runs.
  }
  \label{tab:image_acc}
  \centering
  \scriptsize
  \resizebox{\textwidth}{!}{\begin{tabular}{llcccccccc}
    \hline
    & & \multicolumn{4}{c}{CIFAR-10} & \multicolumn{2}{c}{Flower-102} & \multicolumn{2}{c}{Pet-37} \\
    \textbf{Model} & \textbf{Method} & $N = \textbf{100}$ & \textbf{1000} & \textbf{10000} & \textbf{50000} & \textbf{510} & \textbf{1020} & \textbf{370} & \textbf{3441} \\
    \hline
    \multicolumn{10}{c}{ResNet-50} \\
    \rowcolor{bright-gray} L2-zero & MAP + GS & 67.7 {\tiny(66.0-69.2)} & 87.7 {\tiny(87.2-88.2)} & 94.6 {\tiny(93.8-95.1)} & 97.0 {\tiny(96.9-97.1)} & 86.4 {\tiny(86.0-86.7)} & 92.8 {\tiny(92.4-93.2)} & 87.6 {\tiny(86.7-88.4)} & 93.1 {\tiny(92.9-93.4)} \\
    \rowcolor{bright-gray} & diagEF LA-LML & 67.5 {\tiny(66.6-68.5)} & 88.2 {\tiny(87.8-88.8)} & 95.1 {\tiny(95.1 95.2)} & 96.9 {\tiny(96.9-96.9)} & 85.3 {\tiny(84.7-85.8)} & 92.5 {\tiny(92.4-92.7)} & 88.2 {\tiny(87.8-88.5)} & 93.2 {\tiny(93.0-93.3)} \\
    \rowcolor{bright-gray} & iso DE-ELBO (ours) & 62.3 {\tiny(60.4-63.4)} & 87.1 {\tiny(86.7-87.5)} & 91.3 {\tiny(90.7-91.8)} & 92.9 {\tiny(92.8-93.1)} & 81.6 {\tiny(81.1-82.1)} & 90.0 {\tiny(89.5-90.6)} & 83.2 {\tiny(82.9-83.6)} & 91.8 {\tiny(91.6-92.0)} \\
    L2-SP & MAP + GS & 69.0 {\tiny(65.9-72.3)} & 88.7 {\tiny(88.3-89.4)} & 95.4 {\tiny(95.3-95.6)} & 97.3 {\tiny(97.2-97.4)} & 86.1 {\tiny(85.3-86.7)} & 93.2 {\tiny(93.1-93.3)} & 88.0 {\tiny(87.1-88.8)} & 92.9 {\tiny(92.6-93.1)} \\
    & diagEF LA-LML & 35.0 {\tiny(34.1-36.7)} & 72.0 {\tiny(71.3-72.4)} & 76.4 {\tiny(76.3-76.5)} & 94.1 {\tiny(93.7-94.4)} & 22.9 {\tiny(18.8-29.6)} & 82.7 {\tiny(82.6-82.7)} & 86.7 {\tiny(86.4-87.3)} & 90.6 {\tiny(90.5-90.7)} \\
    & iso DE-ELBO (ours) & 69.9 {\tiny(67.4-71.9)} & 88.5 {\tiny(86.9-89.5)} & 95.1 {\tiny(95.0-95.2)} & 96.8 {\tiny(96.7-96.9)} & 84.5 {\tiny(84.5-84.6)} & 91.8 {\tiny(91.6-92.1)} & 87.5 {\tiny(87.2-87.8)} & 92.9 {\tiny(92.7-93.0)} \\
    \rowcolor{bright-gray} PTYL (SSL) & MAP + GS & 57.5 {\tiny(56.1-58.6)} & 78.4 {\tiny(77.8-79.0)} & 90.6 {\tiny(90.1-90.8)} & 96.6 {\tiny(96.5-96.6)} & 81.9 {\tiny(80.9-82.8)} & 89.7 {\tiny(89.2-90.1)} & 58.3 {\tiny(55.6-60.7)} & 86.3 {\tiny(85.9-86.8)} \\
    \rowcolor{bright-gray} & iso DE-ELBO (ours) & 60.2 {\tiny(59.5-60.7)} & 78.1 {\tiny(77.5-78.8)} & 90.6 {\tiny(90.3-90.8)} & 96.7 {\tiny(96.6-96.7)} & 76.8 {\tiny(76.5-77.2)} & 84.9 {\tiny(84.6-85.1)} & 56.6 {\tiny(56.1-57.2)} & 80.1 {\tiny(80.0-80.3)} \\
    PTYL & MAP + GS & 70.1 {\tiny(69.2-71.4)} & 89.8 {\tiny(89.5-90.3)} & 95.6 {\tiny(95.5-95.8)} & 97.0 {\tiny(96.8-97.2)} & 86.3 {\tiny(85.8-86.6)} & 92.9 {\tiny(92.6-93.1)} & 87.9 {\tiny(87.5-88.2)} & 93.0 {\tiny(92.8-93.2)} \\
    & iso DE-ELBO (ours) & 70.0 {\tiny(67.9-72.1)} & 89.2 {\tiny(89.0-89.5)} & 95.1 {\tiny(94.9-95.4)} & 96.9 {\tiny(96.8-97.0)} & 84.6 {\tiny(84.5-84.6)} & 91.8 {\tiny(91.7-92.0)} & 87.5 {\tiny(87.3-87.8)} & 92.9 {\tiny(92.7-93.0)} \\
    \hline
    \multicolumn{10}{c}{ViT-B/16} \\
    \rowcolor{bright-gray} L2-SP & MAP + GS & 86.5 {\tiny(85.3-87.7)} & 93.8 {\tiny(93.3-94.0)} & 97.2 {\tiny(97.1-97.2)} & 98.2 {\tiny(98.2-98.2)} & 85.8 {\tiny(84.9-86.6)} & 91.4 {\tiny(88.3-93.0)} & 88.1 {\tiny(87.1-89.4)} & 93.3 {\tiny(93.2-93.5)} \\
    \rowcolor{bright-gray} & iso DE-ELBO (ours) & 88.5 {\tiny(87.1-89.8)} & 94.1 {\tiny(94.0-94.2)} & 97.4 {\tiny(97.3-97.6)} & 98.2 {\tiny(98.1-98.3)} & 80.5 {\tiny(78.0-84.0)} & 89.2 {\tiny(88.5-89.6)} & 88.9 {\tiny(88.3-89.5)} & 93.6 {\tiny(93.5-93.7)} \\
    \hline
    \multicolumn{10}{c}{ConvNeXt-Tiny} \\
    \rowcolor{bright-gray} L2-SP & MAP + GS & 85.4 {\tiny(84.3-86.4)} & 94.1 {\tiny(93.9-94.3)} & 96.9 {\tiny(96.7-97.0)} & 97.9 {\tiny(97.8-98.0)} & 88.7 {\tiny(88.3-89.3)} & 94.2 {\tiny(92.3-95.7)} & 89.1 {\tiny(87.4-90.4)} & 94.3 {\tiny(94.2-94.5)} \\
    \rowcolor{bright-gray} & diagEF LA-LML & 80.6 {\tiny(79.8-81.3)} & 88.7 {\tiny(88.5-88.9)} & 92.6 {\tiny(92.5-92.7)} & 95.0 {\tiny(95.0-95.1)} & 79.2 {\tiny(77.8-80.0)} & 92.0 {\tiny(92.0-92.1)} & 90.3 {\tiny(89.7-90.7)} & 93.6 {\tiny(93.6-93.6)} \\
    \rowcolor{bright-gray} & iso DE-ELBO (ours) & 83.9 {\tiny(82.9-85.3)} & 94.5 {\tiny(94.4-94.6)} & 97.1 {\tiny(97.0-97.1)} & 97.8 {\tiny(97.8-97.8)} & 88.0 {\tiny(86.8-89.0)} & 94.1 {\tiny(93.9-94.2)} & 90.7 {\tiny(89.7-91.5)} & 94.3 {\tiny(94.2-94.5)} \\
    \hline
  \end{tabular}}
\end{table*}
\setlength{\tabcolsep}{6pt}

\setlength{\tabcolsep}{2pt}
\begin{table*}[htbp!]
  \caption{NLL on CIFAR-10, Flower-102, and Pet-37 test sets for different probabilistic models, methods, and backbones. We report the mean (min-max) over 3 separately-sampled training sets. MAP + GS requires 24 different SGD runs for L2-zero, 144 for L2-SP, and 240 for PTYL, while diagEF LA-LML~\citep{immer2021scalable} and iso DE-ELBO (ours) require 4 different SGD runs.
  }
  \label{tab:image_nll}
  \centering
  \scriptsize
  \resizebox{\textwidth}{!}{\begin{tabular}{llcccccccc}
    \hline
    & & \multicolumn{4}{c}{CIFAR-10} & \multicolumn{2}{c}{Flower-102} & \multicolumn{2}{c}{Pet-37} \\
    \textbf{Model} & \textbf{Method} & $N = \textbf{100}$ & \textbf{1000} & \textbf{10000} & \textbf{50000} & \textbf{510} & \textbf{1020} & \textbf{370} & \textbf{3441} \\
    \hline
    \multicolumn{10}{c}{ResNet-50} \\
    \rowcolor{bright-gray} L2-zero & MAP + GS & 0.97 {\tiny(0.94-1.03)} & 0.41 {\tiny(0.39-0.44)} & 0.19 {\tiny(0.19-0.20)} & 0.10 {\tiny(0.10-0.10)} & 0.65 {\tiny(0.63-0.68)} & 0.35 {\tiny(0.32-0.38)} & 0.42 {\tiny(0.37-0.47)} & 0.24 {\tiny(0.24-0.25)} \\
    \rowcolor{bright-gray} & diagEF LA-LML & 1.05 {\tiny(1.02-1.06)} & 0.46 {\tiny(0.44-0.48)} & 0.20 {\tiny(0.19-0.21)} & 0.12 {\tiny(0.11-0.12)} & 0.71 {\tiny(0.69-0.73)} & 0.36 {\tiny(0.35-0.37)} & 0.40 {\tiny(0.37-0.42)} & 0.29 {\tiny(0.28-0.29)} \\ 
    \rowcolor{bright-gray} & iso DE-ELBO (ours) & 2.76 {\tiny(2.51-3.18} & 0.54 {\tiny(0.53-0.55)} & 0.39 {\tiny(0.25-0.64)} & 0.28 {\tiny(0.26-0.29)} & 0.84 {\tiny(0.78-0.92)} & 0.44 {\tiny(0.41-0.46)} & 0.80 {\tiny(0.75-0.83)} & 0.27 {\tiny(0.26-0.28)} \\
    L2-SP & MAP + GS & 0.94 {\tiny(0.85-1.02)} & 0.40 {\tiny(0.39-0.41)} & 0.15 {\tiny(0.15-0.16)} & 0.09 {\tiny(0.09-0.09)} & 0.65 {\tiny(0.63-0.68)} & 0.33 {\tiny(0.32-0.35)} & 0.41 {\tiny(0.36-0.45)} & 0.27 {\tiny(0.24-0.30)} \\
    & diagEF LA-LML & 2.30 {\tiny(2.30-2.30)} & 1.18 {\tiny(1.17-1.19)} & 0.78 {\tiny(0.78-0.78)} & 0.18 {\tiny(0.17-0.19)} & 4.62 {\tiny(4.62-4.62)} & 2.60 {\tiny(2.60-2.61)} & 1.34 {\tiny(1.28-1.37)} & 0.41 {\tiny(0.41-0.41)} \\ 
    & iso DE-ELBO (ours) & 1.01 {\tiny(0.92-1.12)} & 0.44 {\tiny(0.42-0.46}) & 0.24 {\tiny(0.23-0.24)} & 0.12 {\tiny(0.12-0.12)} & 0.71 {\tiny(0.67-0.74)} & 0.38 {\tiny(0.38-0.39)} & 0.51 {\tiny(0.49-0.53)} & 0.24 {\tiny(0.24-0.24)} \\
    \rowcolor{bright-gray} PTYL (SSL) & MAP + GS & 1.34 {\tiny(1.20-1.41)} & 0.50 {\tiny(0.49-0.52)} & 0.23 {\tiny(0.22-0.24)} & 0.11 {\tiny(0.11-0.11)} & 0.84 {\tiny(0.81-0.86)} & 0.49 {\tiny(0.46-0.51)} & 1.64 {\tiny(1.55-1.68)} & 0.54 {\tiny(0.50-0.56)} \\
    \rowcolor{bright-gray} & iso DE-ELBO (ours) & 1.36 {\tiny(1.33-1.43)} & 0.76 {\tiny(0.74-0.78)} & 0.29 {\tiny(0.28-0.31)} & 0.12 {\tiny(0.12-0.12)} & 1.10 {\tiny(1.05-1.15)} & 0.72 {\tiny(0.70-0.74)} & 2.07 {\tiny(2.04-2.08)} & 0.68 {\tiny(0.67-0.69)} \\
    PTYL & MAP + GS &  0.90 {\tiny(0.87-0.94)} & 0.36 {\tiny(0.34-0.37)} & 0.15 {\tiny(0.14-0.16)} & 0.10 {\tiny(0.10-0.10)} & 0.65 {\tiny(0.63-0.68)} & 0.33 {\tiny(0.32-0.34)} & 0.42 {\tiny(0.37-0.46)} & 0.26 {\tiny(0.21-0.31)} \\
    & iso DE-ELBO (ours) & 1.00 {\tiny(0.92-1.08)} & 0.46 {\tiny(0.46-0.47)} & 0.23 {\tiny(0.22-0.24)} & 0.12 {\tiny(0.12-0.12)} & 0.71 {\tiny(0.67-0.74)} & 0.38 {\tiny(0.38-0.39)} & 0.51 {\tiny(0.49-0.53)} & 0.24 {\tiny(0.24-0.24)}  \\
    \hline
    \multicolumn{10}{c}{ViT-B/16} \\
    \rowcolor{bright-gray} L2-SP & MAP + GS & 0.46 {\tiny(0.41-0.51)} & 0.25 {\tiny(0.20-0.28)} & 0.10 {\tiny(0.10-0.10)} & 0.06 {\tiny(0.06-0.06)} & 0.69 {\tiny(0.64-0.72)} & 0.42 {\tiny(0.34-0.54)} & 0.42 {\tiny(0.38-0.47)} & 0.25 {\tiny(0.24-0.29)} \\
    \rowcolor{bright-gray} & iso DE-ELBO (ours) & 0.36 {\tiny(0.32-0.41)} & 0.24 {\tiny(0.23-0.26)} & 0.12 {\tiny(0.12-0.13)} & 0.07 {\tiny(0.07-0.07)} & 0.90 {\tiny(0.73-1.03)} & 0.51 {\tiny(0.48-0.55)} & 0.40 {\tiny(0.38-0.42)} & 0.30 {\tiny(0.30-0.31)} \\
    \hline
    \multicolumn{10}{c}{ConvNeXt-Tiny} \\
    \rowcolor{bright-gray} L2-SP & MAP + GS & 0.46 {\tiny(0.44-0.50)} & 0.22 {\tiny(0.19-0.24)} & 0.11 {\tiny(0.10-0.12)} & 0.07 {\tiny(0.07-0.07)} & 0.54 {\tiny(0.51-0.56)} & 0.28 {\tiny(0.23-0.38)} & 0.38 {\tiny(0.30-0.46)} & 0.18 {\tiny(0.18-0.18)} \\
    \rowcolor{bright-gray} & diagEF LA-LML & 1.02 {\tiny(0.94-1.09)} & 0.40 {\tiny(0.40-0.41)} & 0.24 {\tiny(0.24-0.24)} & 0.16 {\tiny(0.16-0.16)} & 2.88 {\tiny(2.87-2.90)} & 0.95 {\tiny(0.94-0.96)} & 0.43 {\tiny(0.43-0.43)} & 0.24 {\tiny(0.24-0.24)} \\
    \rowcolor{bright-gray} & iso DE-ELBO (ours) & 0.54 {\tiny(0.48-0.59)} & 0.23 {\tiny(0.22-0.24)} & 0.12 {\tiny(0.11-0.12)} & 0.07 {\tiny(0.07-0.07)} & 0.58 {\tiny(0.53-0.64)} & 0.30 {\tiny(0.29-0.31)} & 0.32 {\tiny(0.27-0.36)} & 0.22 {\tiny(0.22-0.22)} \\
    \hline
  \end{tabular}}
\end{table*}
\setlength{\tabcolsep}{6pt}

\setlength{\tabcolsep}{2pt}
\setcounter{table}{5}
\begin{table*}[htbp!]
  \caption{ECE on CIFAR-10 test set for different probabilistic models, methods, and backbones. We report the mean (min-max) over 3 separately-sampled training sets. MAP + GS requires 24 different SGD runs for L2-zero, 144 for L2-SP, and 240 for PTYL, while diagEF LA-LML~\citep{immer2021scalable} and iso DE-ELBO (ours) require 4 different SGD runs.
  }
  \label{tab:image_ece}
  \centering
  \scriptsize
  \begin{tabular}{llcccc}
    \hline
    & & \multicolumn{4}{c}{CIFAR-10} \\
    \textbf{Model} & \textbf{Method} & $N = \textbf{100}$ & \textbf{1000} & \textbf{10000} & \textbf{50000} \\
    \hline
    \multicolumn{6}{c}{ResNet-50} \\
    \rowcolor{bright-gray} L2-SP & diagEF LA-LML & 25.0 {\tiny(24.0–26.6)} & 32.4 {\tiny(30.8–33.9)} & 13.5 {\tiny(13.0–13.8)} & 1.8 {\tiny(1.7–1.9)} \\
    \rowcolor{bright-gray} & iso ELBO & 24.9 {\tiny(21.7–28.1)} & 26.1 {\tiny(18.6–38.0)} & 43.5 {\tiny(42.6–45.2)} & 2.3 {\tiny(1.9–2.7)} \\
    \rowcolor{bright-gray} & iso DE-ELBO (ours) & 11.7 {\tiny(10.0–13.2)} & \phantom{0}5.5 {\tiny(\phantom{0}3.6–\phantom{0}6.7)} & \phantom{0}3.3 {\tiny(\phantom{0}3.2–\phantom{0}3.5)} & 1.8 {\tiny(1.7–1.9)} \\
    \multicolumn{6}{c}{ConvNeXt-Tiny} \\
    \rowcolor{bright-gray} L2-SP & diagEF LA-LML & 37.5 {\tiny(34.2–39.9)} & 10.8 {\tiny(10.6–11.1)} & \phantom{0}4.6 {\tiny(\phantom{0}4.5–\phantom{0}4.9)} & 2.2 {\tiny(2.2–2.2)} \\
    \rowcolor{bright-gray} & iso ELBO & 59.7 {\tiny(59.4–60.3)} & 43.9 {\tiny(40.6–47.8)} & 12.1 {\tiny(12.0–12.1)} & 1.6 {\tiny(1.5–1.6)} \\
    \rowcolor{bright-gray} & iso DE-ELBO (ours) & \phantom{0}6.6 {\tiny(\phantom{0}5.6–\phantom{0}7.6)} & \phantom{0}3.5 {\tiny(\phantom{0}3.2–\phantom{0}3.6)} & \phantom{0}1.7 {\tiny(\phantom{0}1.6–\phantom{0}1.8)} & 0.8 {\tiny(0.7–0.8)} \\
    \hline
  \end{tabular}
\end{table*}
\setlength{\tabcolsep}{6pt}

\newpage

\begin{figure}[htbp!]
  \centering
  \includegraphics[width=0.5\linewidth]{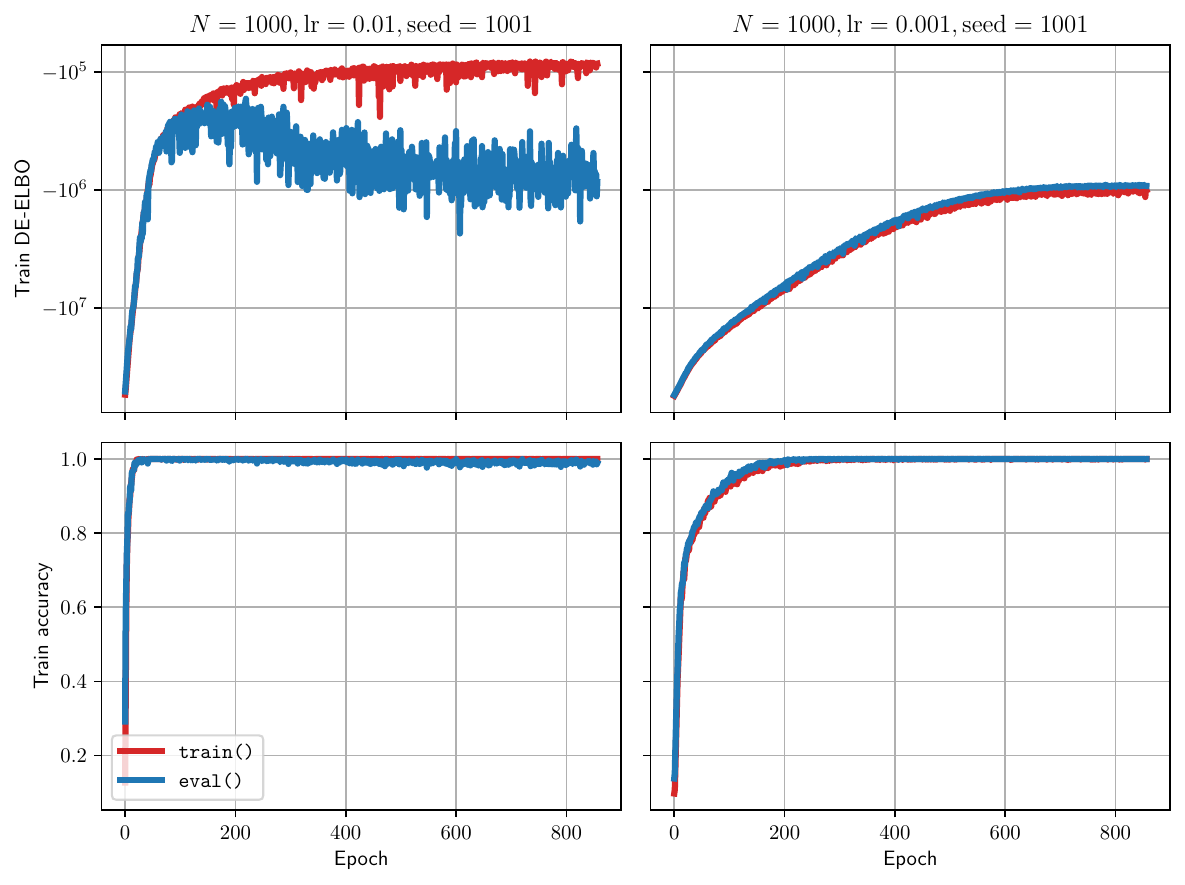}
  \caption{Train set DE-ELBO and accuracy on CIFAR-10 with $N=1000$ using ResNet-50 over epochs for L2-SP. We use just one sample per training step to approximate the expected log-likelihood.}
  \label{fig:train_eval_comparison}
\end{figure}

\textbf{Changes to training performance from running/computed mean and variance in batch normalization layers.}
During training, batch normalization layers keep running estimates of its computed mean and variance, which are then used for normalization during evaluation.
We find that when using batch normalization, the ELBO and accuracy on the train set can change between \texttt{train()} and \texttt{eval()} mode (see Fig.~\ref{fig:train_eval_comparison}).
We recommend using the computed mean and variance to evaluate the ELBO on the training set for model selection since this mode is used for normalization during evaluation.

\setlength{\tabcolsep}{2pt}
\begin{table*}[htbp!]
    \caption{Accuracy on News-4 test set for different probabilistic models, methods, and backbones. We report the mean (min-max) over 3 separately-sampled training sets. MAP + GS requires 24 different SGD runs for L2-zero, 144 for L2-SP, and 240 for PTYL, while diagEF LA-LML~\citep{immer2021scalable} and iso DE-ELBO (ours) require 4 different SGD runs.
    }
    \label{tab:text_acc}
    \centering
    \scriptsize
    \begin{tabular}{llcccc}
        \hline
        & & \multicolumn{4}{c}{News-4} \\
        \textbf{Model} & \textbf{Method} & $N = \textbf{400}$ & \textbf{4000} & \textbf{40000} & \textbf{120000} \\
        \hline
        \multicolumn{6}{c}{BERT-base} \\
        \rowcolor{bright-gray} L2-SP & MAP + GS & 85.1 {\tiny(83.8-87.0)} & 89.4 {\tiny(89.3-89.6)} & 92.4 {\tiny(91.9-92.8)} & 93.4	{\tiny(92.2-94.0)} \\
        \rowcolor{bright-gray} & diagEF LA-LML & 84.2 {\tiny(83.5-85.3)} & 89.8 {\tiny(89.7-90.0)} & 93.1 {\tiny(93.0-93.4)} & 92.5 {\tiny(92.5-92.6)} \\
        \rowcolor{bright-gray} & iso DE-ELBO (ours) & 87.8 {\tiny(87.4-88.2)} & 91.1 {\tiny(90.9-91.2)} & 93.5 {\tiny(93.4-93.6)} & 94.4 {\tiny(94.3-94.5)} \\
        \hline
    \end{tabular}
    \end{table*}
\setlength{\tabcolsep}{6pt}

\setlength{\tabcolsep}{2pt}
\begin{table*}[htbp!]
    \caption{NLL on News-4 test set for different probabilistic models, methods, and backbones. We report the mean (min-max) over 3 separately-sampled training sets. MAP + GS requires 24 different SGD runs for L2-zero, 144 for L2-SP, and 240 for PTYL, while diagEF LA-LML~\citep{immer2021scalable} and iso DE-ELBO (ours) require 4 different SGD runs.
    }
    \label{tab:text_nll}
    \centering
    \scriptsize
    \begin{tabular}{llcccc}
        \hline
        & & \multicolumn{4}{c}{News-4} \\
        \textbf{Model} & \textbf{Method} & $N = \textbf{400}$ & \textbf{4000} & \textbf{40000} & \textbf{120000} \\
        \hline
        \multicolumn{6}{c}{BERT-base} \\
        \rowcolor{bright-gray} L2-SP & MAP + GS & 0.09 {\tiny(0.05-0.17)} & 0.03 {\tiny(0.03-0.03)} & 0.02 {\tiny(0.02-0.02)} & 0.02 {\tiny(0.02-0.02)} \\
        \rowcolor{bright-gray} & diagEF LA-LML & 0.08 {\tiny(0.08-0.09)} & 0.03 {\tiny(0.03-0.03)} & 0.02 {\tiny(0.02-0.02)} & 0.02 {\tiny(0.02-0.02)} \\
        \rowcolor{bright-gray} & iso DE-ELBO (ours) & 0.10 {\tiny(0.10-0.11)} & 0.08 {\tiny(0.08-0.09)} & 0.03 {\tiny(0.03-0.03)} & 0.02 {\tiny(0.02-0.02)} \\
        \hline
    \end{tabular}
    \end{table*}
\setlength{\tabcolsep}{6pt}

\subsection{Importance Weighted ELBO for Model B}

For transfer learning of image classifiers from Case Study B, the results in Sec.~\ref{sec:caseB} show that our \emph{data-emphasized ELBO} outperforms the standard ELBO, favoring settings of $\psi, \eta$ that produce higher test accuracy.
Our hypothesis about \emph{why} this occurs is that in our chosen target scenario, the ELBO objective itself is too loose a bound on the log marginal likelihood for reliable selection, whereas setting $\kappa=D/N$ helps account for the misspecified isotropic $q$ to deliver better fitting models.

Recall the bounding relation between the exact log marginal likelihood and the ELBO:
\begin{align}
     \underbrace{\log \mathbb{E}_{q_\psi(\theta)} \left[ \frac{p_{\eta}(y_{1:N} | \theta) p_{\eta}(\theta)}{q_\psi(\theta)} \right]}_{\log p_{\eta}(y_{1:N})} 
     \geq
     \underbrace{\mathbb{E}_{q_\psi(\theta)} \left[ \log \frac{p_{\eta}(y_{1:N} | \theta) p_{\eta}(\theta)}{q_\psi(\theta)} \right]}_{J_{\text{ELBO}}(y_{1:N}, \psi, \eta)}
\end{align}

Naturally, if we could find tighter bounds than the ELBO we might hope these could be used for improved practical model selection, delivering good predictive accuracy even when $q$ remains misspecified as isotropic compared to an ideal full-rank covariance.
One possible candidate for a tighter bound than the ELBO is the \emph{importance weighted ELBO} (IWELBO) objective of \citep{burda2016importance}. \begin{align}
    J_{\text{IWELBO}} := \mathbb{E}_{\theta_1, \dots, \theta_S \sim q_\psi(\theta)} \left[ \log \frac{1}{S} \sum_{s=1}^S \exp \left( \sum_{i=1}^{N} \log p_{\eta}(y_i | \theta_s) + \log p_{\eta}(\theta_s) - \log q_\psi(\theta_s) \right) \right]
\end{align}
\citeauthor{burda2016importance} show that the IWELBO is a better estimator of the log of the evidence than the ELBO, with estimation quality improving as the number of samples $S$ increases. 
\citeauthor{burda2016importance}'s Theorem 1 says that, if $\frac{p(y_{1:N}, \theta)}{q_\psi(\theta)}$ is bounded, then in the limit as $S \rightarrow \infty$, the IWELBO will converge to the log marginal likelihood.

We can also introduce a \emph{data-emphasized IWELBO} (DE-IWELBO)
\begin{align}
    J_{\text{DE-IWELBO}} := \mathbb{E}_{\theta_1, \dots, \theta_S \sim q_\psi(\theta)} \left[ \log \frac{1}{S} \sum_{s=1}^S \exp \left( \kappa \cdot \sum_{i=1}^{N} \log p_{\eta}(y_i | \theta_s) + \log p_{\eta}(\theta_s) - \log q_\psi(\theta_s) \right) \right]
\end{align}
For simplicity, we again set $\kappa = \frac{D}{N}$ like with our DE-ELBO. Future work could investigate theoretical grounding for how to set $\kappa$ in the DE-IWELBO.

In principle, a training algorithm that estimates $\psi,\eta$ that maximize IWELBO with very large $S$ should enjoy the Ockham's razor benefits of the true marginal likelihood.
However, for large DNNs using very large $S$ requires $S$ forward passes at every input image and would be prohibitively expensive.

To understand if the IWELBO offers a \emph{practical} route forward, we thus experiment with whether the $S=50$ IWELBO could resolve the model selection issues seen in Fig.~\ref{fig:elbo_comparison}. 
We use the same pink and purple $\psi$ from Fig.~\ref{fig:elbo_comparison}. We then evaluate the $S=50$ IWELBO and $S=50$ DE-IWELBO at each $\psi$ across a range of $\eta$. 

Results are shown in the figure below. 
While we verified the IWELBO yields numerically higher values than the ELBO at each $\psi,\eta$, the differences are slight compared to the overall y-axis scale. The overall trends and takeaways for DE-IWELBO vs IWELBO match the takeaways from Fig.~\ref{fig:elbo_comparison} for DE-ELBO vs ELBO.

Even with $S=50$ samples, the IWELBO favors the low-test-accuracy $\psi$ over the high-test-accuracy $\psi$.
This suggests that \citeauthor{burda2016importance}'s tighter objective isn't an immediate resolution to the problems with ELBO model selection raised in Contribution 1, at least with $S=50$.
Interestingly, the DE-IWELBO with $S = 50$ favors $q$ that produce higher test accuracy.


\begin{figure}[htbp!]
  \centering
  \includegraphics[width=0.5\linewidth]{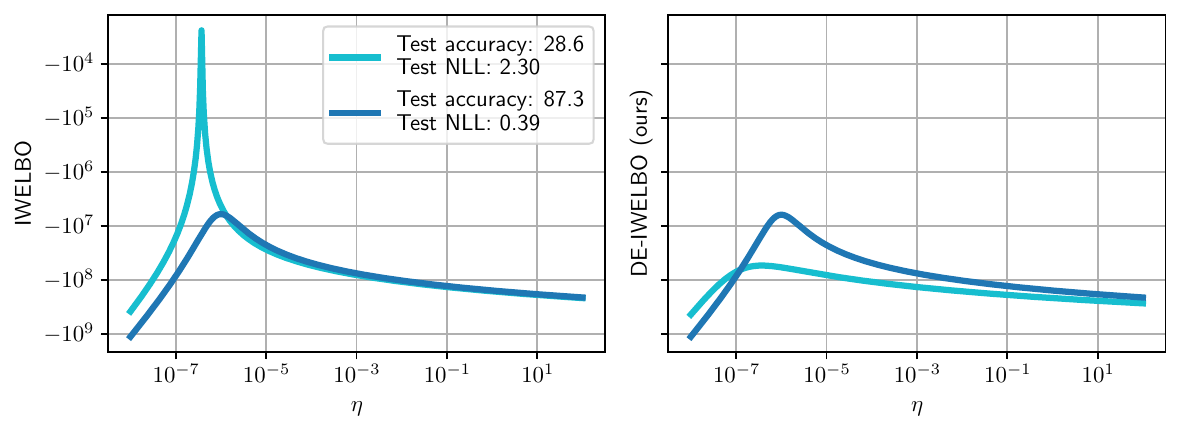}
  \caption{Model selection comparison for the $S=50$ IWELBO and $S=50$ DE-IWELBO across a range of hyperparameters $\eta = \lambda = \tau$ for Model B in Eq.~\eqref{eq:caseB_joint_pdf}.
  Task: ResNet-50 fine-tuned on CIFAR-10 with $N=1000$. 
  The light blue curve uses the same low-test-accuracy $\psi$ value as in Fig.~\ref{fig:elbo_comparison}; the dark blue curve uses the same high-test-accuracy $\psi$ value as in Fig.~\ref{fig:elbo_comparison}.
  \textbf{Takeaway: $S=50$ is near the limit of practical affordability for large DNNs, and even here the tighter IWELBO is not enough to overcome the underfitting of the standard ELBO raised in Fig.~\ref{fig:elbo_comparison}.
  }
  }
  \label{fig:iwelbo_comparison}
\end{figure}

\end{document}